\ifdefined\pdfminorversion\pdfminorversion=7\fi
\documentclass[10pt,twocolumn,letterpaper]{article}
\usepackage[pagenumbers]{cvpr}

\usepackage{cuted}

\usepackage[expansion=false]{microtype}

\usepackage{placeins}
\usepackage{multirow}

\usepackage{colortbl}

\usepackage[T1]{fontenc}

\AtBeginDocument{%
  \setlength{\abovedisplayskip}{6pt plus 2pt minus 2pt}%
  \setlength{\belowdisplayskip}{6pt plus 2pt minus 2pt}%
  \setlength{\abovedisplayshortskip}{0pt plus 2pt}%
  \setlength{\belowdisplayshortskip}{4pt plus 2pt minus 2pt}%
}

\usepackage{float,multicol}
\definecolor{cvprblue}{rgb}{0.21,0.49,0.74}
\usepackage[breaklinks,colorlinks,allcolors=cvprblue]{hyperref}

\title{Particle Forcing: Guiding Spatio-Temporal Layouts, \\Learning Physics-Aware Generation}
\author{%
  Liang Dong\textsuperscript{1,2}\quad
  Haowen Xu\textsuperscript{1}\quad
  Tieyue Yin\textsuperscript{1,3}\quad
  Yunfeng Wu\textsuperscript{1,4}\quad
  Jun Sun\textsuperscript{3}\\
  Songhua Liu\textsuperscript{1,*}\quad
  Yingtian Zou\textsuperscript{1,*}\\[3pt]
  {\normalsize
  \textsuperscript{1}Shanghai Jiao Tong University\quad
  \textsuperscript{2}Carnegie Mellon University}\\
  {\normalsize
  \textsuperscript{3}Nanjing University\quad
  \textsuperscript{4}The University of Hong Kong}\\[2pt]
  {\small \textsuperscript{*}Corresponding authors}
}
\hypersetup{pdfauthor={Liang Dong, Haowen Xu, Tieyue Yin, Yunfeng Wu, Jun Sun, Songhua Liu, Yingtian Zou}}
\begin{document}
\maketitle
\setlength{\stripsep}{8pt}
\begin{strip}
    \centering
    \includegraphics[width=0.95\textwidth]{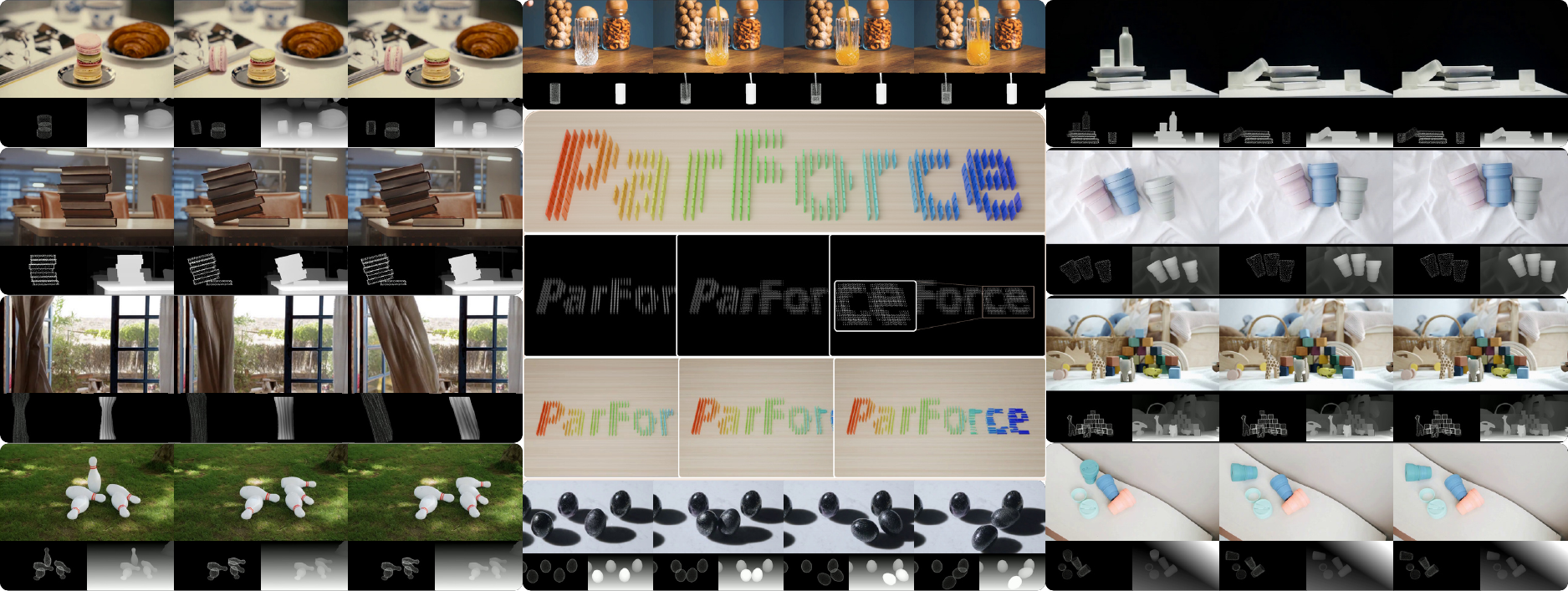}
    \captionsetup{hypcap=false}
    \captionof{figure}{Physics-aware video generation with \textbf{ParForce}. It guides spatio-temporal scene evolution through perspective-aware particle representations and surface depth constraints, enabling controllable and physically plausible video generation across diverse scenarios. Each sequence progresses from left to right, with the corresponding particle and depth conditions visualized beneath the generated frames.}
    \captionsetup{hypcap=true}
    \label{fig:teaser}

    \begin{minipage}{\textwidth}
      \vspace{7pt}
      \setlength{\multicolsep}{0pt}
      \renewenvironment{abstract}{%
        \begin{multicols}{2}[\hbox to \dimexpr(\linewidth-\columnsep)/2\relax{\hss\large\bfseries Abstract\hss}\vspace{7pt}]%
        \itshape\noindent\ignorespaces
      }{\par\end{multicols}}%
      \begin{abstract}
Generating physically coherent videos from a first frame, a scene description, and an action instruction requires coordinating subject motion with scene geometry and visibility over time. Existing methods commonly use image-plane motion, trajectories, geometry, or rendered outputs as control conditions. Three-dimensional spatial information plays a crucial role in guiding physically plausible video generation, as it helps resolve the inherent ambiguity of 2D visual conditions by providing essential cues about scene geometry, depth ordering, and occlusion relationships. We therefore introduce \textbf{ParForce}, a physics-conditioned video generation framework that jointly guides spatial-temporal layouts using perspective particle state and a surface-depth condition. Obtaining fully paired supervision from real-world videos is inherently challenging, as monocular observations cannot uniquely recover the underlying 3D dynamics, particularly in regions involving \mbox{occlusions and complex physical interactions.}
\par\columnbreak\noindent To address this limitation, we construct scalable synthetic video--particle--depth triplets that provide spatially and temporally aligned motion and geometric supervision. We adopt a hybrid training strategy that incorporates real-world video--depth pairs to bridge the synthetic-to-real domain gap, enabling physics-aware video generation without requiring particle annotations from real-world scenes. Extensive experiments across three video generation backbones demonstrate the effectiveness and generalizability of ParForce, achieving up to 74.4\% lower FVD and 57.1\% lower optical-flow Endpoint Error (EPE) compared with existing motion-conditioned approaches under matched backbone settings. Furthermore, we distill the particle-based dynamics modeling into a student model, enabling video generation from a first frame and text without explicitly estimating particle states at inference time and improving the \mbox{Physics-IQ Verified score by 7.36 points.}
\end{abstract}

    \end{minipage}
\end{strip}
\clearpage
{\linespread{0.96}\selectfont\section{Introduction}
\label{sec:intro}

Video diffusion models have made substantial progress in visual quality,
temporal coherence, and alignment with text prompts
\cite{wan2025wanopenadvancedlargescale,kong2025hunyuanvideosystematicframeworklarge,yang2025cogvideoxtexttovideodiffusionmodels}.
However, for image-to-video generation driven by a first frame, a scene
description, and a natural-language action instruction, the challenge
extends beyond producing coherent frames: the model must predict how a
scene responds to an action over time. This prediction is inherently
decided by three-dimensional spatial relations, as object motion,
contact, and occlusion depend on the geometry of the object and its
surroundings, while both the generator output and common control
conditions lie in the two-dimensional image plane. This mismatch between
two-dimensional representations and three-dimensional physical relations
raises a central question: \textit{How can three-dimensional physical
information be encoded in two-dimensional conditions that can be used by a video
generator?}

Existing approaches condition video generation on explicit motion or physical representations. Go-with-the-Flow uses
optical flow, while Tora and Wan-Move use point trajectories in the image
plane
\cite{burgert2025gowiththeflowmotioncontrollablevideodiffusion,zhang2025toratrajectoryorienteddiffusiontransformer,chu2025wanmovemotioncontrollablevideogeneration}.
Go-with-the-Track uses reference-anchored point tracks and can construct
projected tracks and visibility cues through 3D reconstruction or dynamic
3D tracking~\cite{namekata2026gowiththetrackvideocompositingmotion}.
Simulation-driven methods such as PhysGen and RealWonder compute action
consequences before converting them into video-generation guidance
\cite{liu2024physgenrigidbodyphysicsgroundedimagetovideo,liu2026realwonderrealtimephysicalactionconditioned};
RealWonder, for example, renders optical flow and coarse RGB previews.
Yet similar projected motion can arise from different 3D dynamics.
Flow and projected tracks, even when constructed through 3D preprocessing,
do not explicitly describe the evolving spatial extent of a subject together with its depth ordering relative to surrounding surfaces by themselves.

\begin{figure*}[t]
    \centering
    \includegraphics[width=0.90\textwidth,keepaspectratio]{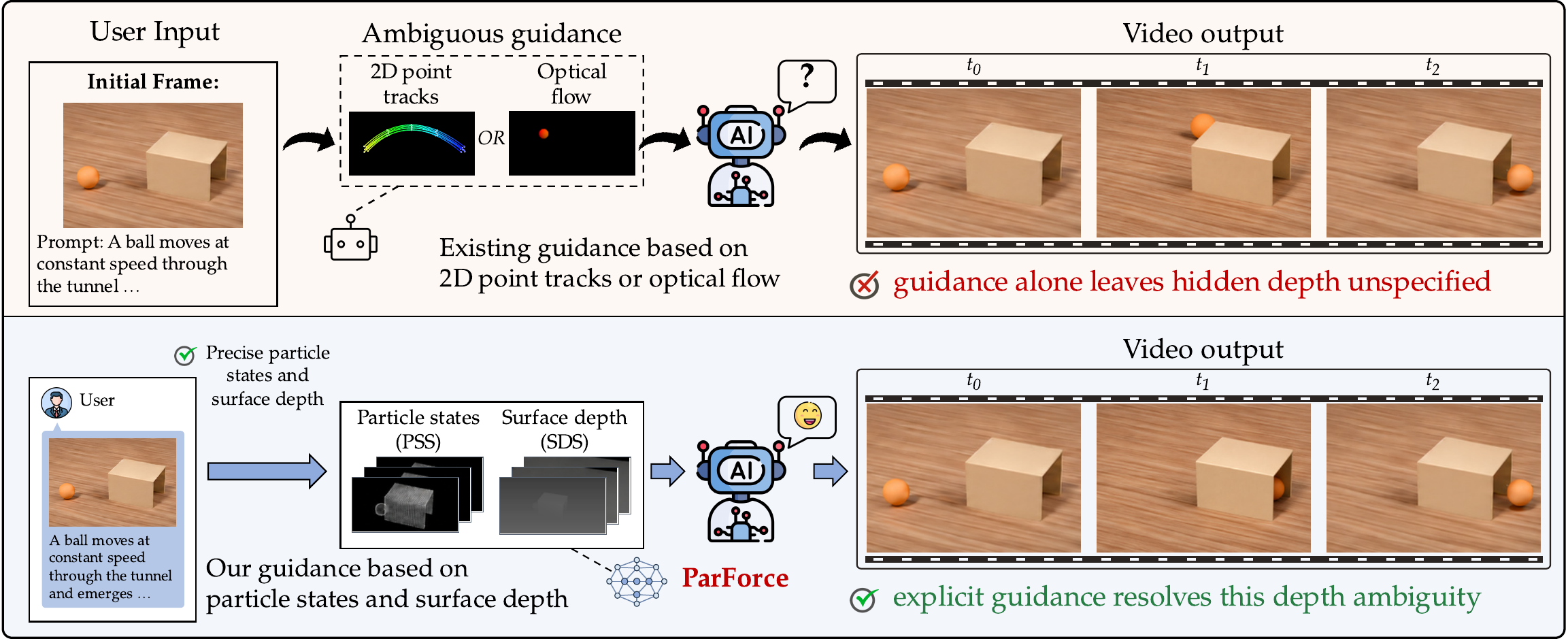}
    \caption{Ambiguity of planar guidance at an occlusion. A perspective particle state sequence encodes the dynamic subject; surface depth encodes scene geometry.}
    \label{fig:introduction}
\end{figure*}


Consider the instruction in Fig.~\ref{fig:introduction}:
\textit{``A ball moves at constant speed through the tunnel.''}
From the input viewpoint, a ball passing through the tunnel and one
passing behind it can follow similar 2D paths. Although the text specifies
the intended passage, flow or projected tracks alone do not explicitly
encode the ball's depth relative to the tunnel during occlusion.
To provide this spatial information, we project simulated 3D particles
into the image plane under the camera
perspective. The resulting \emph{perspective particle state
sequence (PSS)} describes the evolving spatial distribution of the
moving subject, including portions occluded by scene surfaces.

However, PSS alone does not explicitly delineate continuous visible
surfaces. When multiple subjects overlap in projection or pass behind
scene geometry, their particle states retain sampled depths but do not
directly specify surface coverage and occlusion boundaries. We therefore
introduce a \emph{surface-depth sequence (SDS)} to describe the visible
scene geometry. PSS provides the dynamic subject state over time,
including depth cues for occluded portions, while SDS characterizes
continuous scene surfaces and their visibility boundaries. Together,
they resolve depth and occlusion ambiguities under projected overlap.
Combining these conditions, we introduce \textit{ParForce}, a
physics-conditioned framework that guides the spatio-temporal layout of
generated videos. To make these physical conditions available from
practical inputs, we further introduce an augmented physical condition
constructor. Given a first frame and an instruction, it automatically
produces PSS and SDS through scene reconstruction, action simulation,
and rendering.

Training with these conditions requires RGB videos paired with
temporally aligned particle states and surface depth. However, real RGB videos
generally do not uniquely determine dynamic 3D states or occluded
geometry, making temporally aligned PSS--SDS supervision difficult to
acquire and validate at scale. We therefore construct simulated
RGB--PSS--SDS triplets from shared scene geometry and motion. However,
simulated appearance and scene statistics differ from
those of real videos, introducing a sim-to-real gap. To address this gap,
we combine simulated triplets with real RGB--depth pairs during training.
Simulation provides aligned motion and geometry supervision, while real
videos supply natural appearance and dynamics without requiring particle
annotations. The particle condition is disabled for real samples.

We empirically find that the knowledge learned by \textit{ParForce} can
be transferred to a standard image-to-video student through our
\emph{Asymmetric Physics-Guided Distribution Matching Distillation (AP-DMD)}
\cite{yin2024onestepdiffusiondistributionmatching,yin2024improveddistributionmatchingdistillation}.
The physics-conditioned generator serves as the teacher, allowing the
student to internalize explicit physical guidance as an implicit
generative prior. At inference, the four-step student takes only the
first frame and text, eliminating the need to prepare particle or depth
conditions.

Our main contributions are as follows:
\begin{itemize}
    \item We introduce \textbf{\textit{ParForce}}, which uses a
    perspective particle state sequence and a surface-depth sequence as
    complementary physical conditions to jointly resolve spatial and
    occlusion ambiguities under projected motion guidance. ParForce
    improves motion following, geometric consistency, and visual fidelity
    across multiple video generation backbones.

    \item Since real videos paired with particle conditions are difficult
    to obtain, we design a \textbf{\textit{hybrid training strategy}} that
    combines \textit{simulated} video with particle and depth conditions and \textit{real}
    videos with depth conditions only, narrowing the sim-to-real gap
    without requiring real-world particle annotations.

    \item We propose \textbf{\textit{Asymmetric Physics-Guided Distribution
    Matching Distillation (AP-DMD)}}, which internalizes the teacher's
    explicit physical guidance in a four-step student. Using only the
    first frame and text, the student more accurately predicts the
    physical evolution of a scene without explicit physical conditions.
    
\end{itemize}
\par}
\section{Related Work}
\label{sec:related}
\subsection{Physics-Conditioned Video Generation}

Generating physically plausible videos from an initial image and an action description remains challenging because action text alone does not fully specify how objects move, deform, become occluded, or interact with their surroundings. Accordingly, controllable video generation methods introduce explicit motion conditions. Tora encodes user-specified trajectories into spatio-temporal motion features~\cite{zhang2025toratrajectoryorienteddiffusiontransformer}; Go-with-the-Flow controls motion through optical-flow-warped noise~\cite{burgert2025gowiththeflowmotioncontrollablevideodiffusion}; and Go-with-the-Track uses point tracks for motion control and reference-content compositing~\cite{namekata2026gowiththetrackvideocompositingmotion}. A complementary class of physics-grounded methods reconstructs a scene, simulates actions, and converts simulated dynamics into conditions for video generation~\cite{liu2024physgenrigidbodyphysicsgroundedimagetovideo,tan2024physmotionphysicsgroundeddynamicssingle,li2025wonderplaydynamic3dscene,liu2026realwonderrealtimephysicalactionconditioned}.
These efforts show that explicit conditions improve motion control, but different representations expose different physical information to the generator. In particular, the same image-plane trajectory or flow can correspond to different three-dimensional states and depth orderings~\cite{gu2025diffusionshader3dawarevideo}. Reliably reasoning about occlusion, contact, and deformation requires both the evolving state of a dynamic subject and its surface-geometric relation to the surrounding scene. For this reason, ParForce uses PSS and SDS as complementary physical conditions to guide video generation.

\subsection{Diffusion Distillation for Fast Video Generation}

Diffusion distillation compresses a multi-step sampler into a model requiring substantially fewer denoising steps. Progressive distillation and consistency-based methods approximate a teacher's denoising trajectory or consistency mapping~\cite{salimans2022progressivedistillationfastsampling,song2023consistencymodels,feng2026parametrictrajectorydistillationfewstep}. 
Distribution matching distillation (DMD) instead aligns teacher and student output distributions using score or denoising estimates, without requiring the student to reproduce a fixed teacher trajectory~\cite{yin2024onestepdiffusiondistributionmatching,yin2024improveddistributionmatchingdistillation,zhou2024scoreidentitydistillationexponentially}. Existing distillation approaches primarily target sampling acceleration and therefore commonly preserve the same text, image, or control inputs for teacher and student. Physical conditions can also be constructed at inference, but doing so incurs additional reconstruction and simulation costs.
We adopt an asymmetric DMD setup in which a strongly physics-conditioned teacher supervises a student receiving only the first frame and text, transferring explicit physical guidance into an implicit generative prior~\cite{lopezpaz2016unifyingdistillationprivilegedinformation,gupta2015crossmodaldistillationsupervision}.

\raggedbottom
\section{Method}
\label{sec:method}

\subsection{Problem Formulation and Overview}
\label{sec:formulation}

\begin{figure*}[t]
    \centering
    \includegraphics[width=0.985\textwidth]{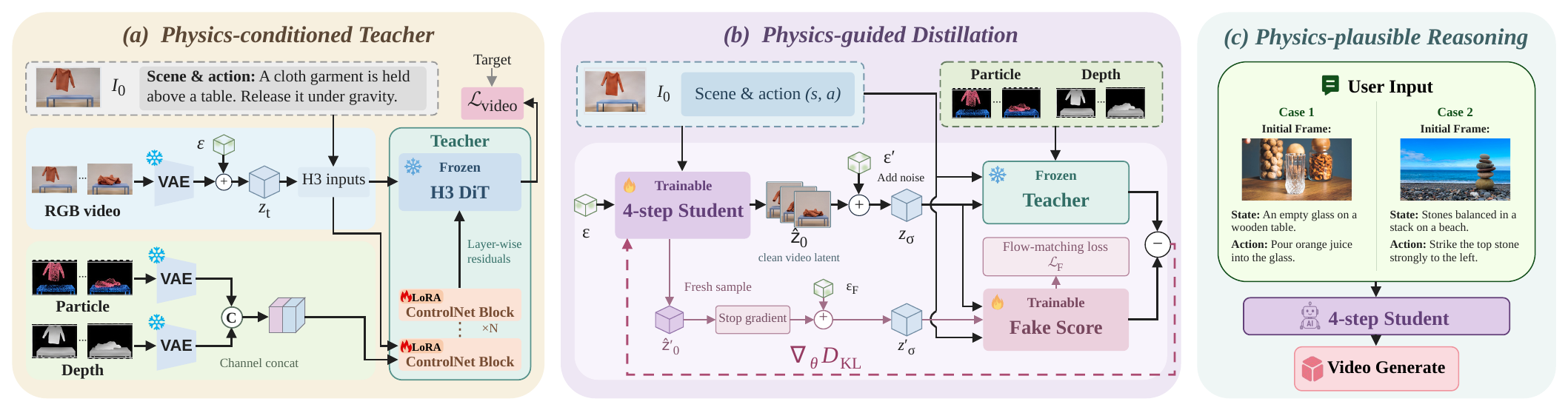}
    \caption{\textbf{ParForce overview.} (a) A perspective particle state sequence and a surface-depth sequence condition the teacher through ControlNet. (b) Asymmetric Physics-Guided Distribution Matching Distillation trains a four-step student. (c) The student generates videos from the initial image, scene description, and action instruction without simulation.}
    \label{fig:parforce-method}
\end{figure*}

Given an initial image $I_0$, an initial-scene description $s$, and an action instruction $a$, we aim to generate a video $\mathbf V$ that preserves the scene's appearance and responds to the action with physically plausible dynamics. The description $s$ specifies the objects, layout, materials, and known initial states, while $a$ specifies the target, manner, direction, relative strength, or ordering of an intervention. Neither description states the resulting motion or final outcome.

ParForce combines physics-conditioned video generation with physics-guided distillation. With $c=(I_0,s,a)$, its physics-conditioned generator and simulator-free student model
\begin{equation}
    p_\phi(\mathbf V\mid c,\mathbf P,\mathbf D),
    \qquad q_\theta(\mathbf V\mid c),
    \label{eq:teacher_student}
\end{equation}
where $\mathbf P$ and $\mathbf D$ denote the video-aligned PSS and SDS, respectively. The generator uses these conditions to guide future spatial layouts and motion. Distillation transfers this physical guidance to a student that receives only the first image and text.

For end-to-end conditioned generation, an \emph{augmented physical condition constructor} reconstructs the initial scene, simulates the instructed action, and renders PSS and SDS. The generator then combines these conditions with the input appearance to generate RGB video. We train it on offline paired data without backpropagating through the constructor. Student inference retains only the video generator.

\subsection{PSS and SDS Conditions}
\label{sec:physical_conditions}

Similar image-plane trajectories can arise from different three-dimensional motions. For example, an object passing in front of or behind an obstacle can follow a similar projected path but exhibit different visibility and occlusion timing. We represent these distinctions through PSS and SDS conditions.

At video time $k$, let $\{\mathbf X_{k,i}\}$ denote three-dimensional subject points and $\pi_C$ the camera projection. Their image positions and intensities are
\begin{equation}
    \mathbf u_{k,i}=\pi_C(\mathbf X_{k,i}),
    \qquad b_{k,i}=g(Z_{k,i}),
    \label{eq:particle_projection}
\end{equation}
where $Z_{k,i}$ is camera-space depth and $g$ is a fixed mapping over the entire video, with nearer points brighter and farther points darker. We render small disks on a black background, retaining sampled points on back-facing or occluded parts of the subject without surface-visibility culling. Points are drawn from far to near where projections overlap, allowing PSS to describe subject motion during occlusion.

Points are sampled in a reference state and remain attached to the subject throughout the sequence. Rigid-body points are bound to local coordinates and follow the body's transformations. Deformable-body points are bound to material-mesh triangles through barycentric coordinates and follow the mesh deformation. These bindings preserve temporal correspondence.

PSS describes discrete subject states but does not fully specify continuous surfaces or visibility boundaries. We therefore provide a separate SDS, denoted $\mathbf D$, for the dynamic subjects and surrounding environment. Depth is rendered independently from scene surfaces for simulated videos and estimated from source images for real videos. Both modalities share the same timeline, camera, and spatial preprocessing, jointly conditioning generation on subject motion and scene geometry.

\subsection{Augmented Physical Condition Constructor}
\label{sec:condition_constructor}

The augmented physical condition constructor uses a vision-language model and scene-specific tools to estimate geometry, camera, materials, and action controls from the initial image and text. It simulates the instructed action and renders PSS and SDS from the same scene states. It uses no future reference frames or trajectories, records assumptions about unobserved properties, and checks initial alignment, support, and contact before passing conditions to the generator. This external tool process is not differentiated through; implementation details are provided in the appendix.

\subsection{Physics-Conditioned Video Generation}
\label{sec:teacher}

\paragraph{Training data.}
We jointly train on simulated RGB/particle/depth triplets and real RGB/depth pairs. Simulation supplies aligned PSS and SDS conditions, while real videos provide natural appearance and depth supervision without requiring real-world particle annotations. No real RGB/particle examples are used. For depth-only real samples, the PSS condition is set to zero in latent space. Each training batch contains equal numbers of simulated and real samples.

\paragraph{Condition injection.}
We use MiniMax-H3~\cite{minimax_h3_release} as the main generative backbone and a ControlNet~\cite{zhang2023controlnet,h3_fun_union2_release} branch to incorporate the physical conditions. A frozen VAE encodes PSS and SDS separately; their latents are concatenated along the feature dimension and fed to the control branch, whose projected residuals are injected at selected backbone layers.

We compared channel fusion, token fusion, and ControlNet for injecting PSS and SDS. ControlNet encodes the conditions in a separate branch and injects residuals at selected frozen H3 layers, giving the most stable mixed simulation--real adaptation (see supplement).

The text condition contains an \emph{initial-scene description} $s$ and an \emph{external-force description} $a$, encoded together with $I_0$ using the frozen image--text encoder. The resulting condition tokens interact with video-latent tokens through H3's joint self-attention at every denoising step. The first image is also VAE-encoded and supplied as a known-frame latent prefix.

\paragraph{Training objective.}
The original H3 backbone, image--text encoder, and VAE remain frozen. We train LoRA~\cite{hu2022lora} parameters in the control branch's attention and feed-forward layers, along with its input and residual projections. At noise level $\sigma\in(0,1)$, we perturb the reference RGB latent $z_{\mathrm{GT}}$:
\begin{equation}
    x_\sigma=(1-\sigma)z_{\mathrm{GT}}+\sigma\epsilon,
    \qquad \epsilon\sim\mathcal N(0,I).
    \label{eq:teacher_noising}
\end{equation}
Under H3's clean-minus-noise convention, the target velocity is $z_{\mathrm{GT}}-\epsilon$, giving
\begin{equation}
    \mathcal L_T=\mathbb E\!\left[
        \operatorname{MSE}\!\left(
            v_\phi(x_\sigma,\sigma;c,\mathbf P,\mathbf D),
            z_{\mathrm{GT}}-\epsilon
        \right)\right].
    \label{eq:teacher_loss}
\end{equation}
The loss covers only the generated video tokens, excluding the known-frame prefix. At inference, the generator produces RGB video from noise and the supplied conditions; future reference RGB is not a generation input.

\begin{table*}[!tp]
\centering
\small
\setlength{\tabcolsep}{4pt}
\renewcommand{\arraystretch}{1.12}
\resizebox{\textwidth}{!}{%
\begin{tabular}{>{\centering\arraybackslash}p{0.17\textwidth}>{\centering\arraybackslash}p{0.12\textwidth}cccccccccc}
\toprule
\multirow{2}{*}{Method} & \multirow{2}{*}{Backbone} & \multicolumn{2}{c}{Visual fidelity} & \multicolumn{3}{c}{Reconstruction} & Motion & \multicolumn{2}{c}{Geometry} & \multicolumn{2}{c}{VLM quality} \\
\cmidrule(lr){3-4}\cmidrule(lr){5-7}\cmidrule(lr){8-8}\cmidrule(lr){9-10}\cmidrule(lr){11-12}
 & & FID $\downarrow$ & FVD $\downarrow$ & LPIPS $\downarrow$ & PSNR $\uparrow$ & SSIM $\uparrow$ & EPE $\downarrow$ & Canny F1 $\uparrow$ & D-nRMSE $\downarrow$ & V-Int. $\uparrow$ & P-Phys. $\uparrow$ \\
\midrule
Tora~\cite{zhang2025toratrajectoryorienteddiffusiontransformer} & \mbox{CogVideoX 5B} & 45.05 & 424.8 & 0.166 & 26.22 & 0.914 & 6.443 & 0.653 & 0.089 & 0.413 & 0.213 \\
Go-with-the-Flow~\cite{burgert2025gowiththeflowmotioncontrollablevideodiffusion} & \mbox{CogVideoX 5B} & 46.62 & 309.1 & 0.133 & 26.20 & 0.931 & 6.971 & 0.709 & 0.075 & 0.373 & 0.133 \\
\rowcolor{gray!10}
Ours & \mbox{CogVideoX 5B} & \textbf{29.31} & \textbf{143.0} & \textbf{0.057} & \textbf{31.04} & \textbf{0.960} & \textbf{3.962} & \textbf{0.823} & \textbf{0.045} & \textbf{0.667} & \textbf{0.293} \\
\specialrule{0.8pt}{1.5pt}{1.5pt}
RealWonder~\cite{liu2026realwonderrealtimephysicalactionconditioned} & \mbox{Wan2.1 1.3B} & 35.47 & 252.2 & 0.082 & 29.37 & 0.946 & 5.638 & 0.759 & 0.052 & 0.407 & 0.187 \\
\rowcolor{gray!10}
Ours & \mbox{Wan2.1 1.3B} & \textbf{20.51} & \textbf{64.56} & \textbf{0.027} & \textbf{35.37} & \textbf{0.974} & \textbf{2.417} & \textbf{0.929} & \textbf{0.030} & \textbf{0.800} & \textbf{0.300} \\
\specialrule{0.8pt}{1.5pt}{1.5pt}
Go-with-the-Track~\cite{namekata2026gowiththetrackvideocompositingmotion} & \mbox{Wan2.2 14B} & \textbf{18.81} & 70.18 & 0.029 & 34.00 & 0.975 & 2.251 & 0.922 & \textbf{0.028} & 0.820 & 0.313 \\
\rowcolor{gray!10}
Ours & \mbox{Wan2.2 14B} & 19.93 & \textbf{62.00} & \textbf{0.025} & \textbf{36.55} & \textbf{0.976} & \textbf{2.238} & \textbf{0.936} & 0.029 & \textbf{0.860} & \textbf{0.327} \\
\specialrule{0.8pt}{1.5pt}{1.5pt}
Original H3 & \mbox{H3 33B} & 34.39 & 290.7 & 0.141 & 27.25 & 0.932 & 8.826 & 0.677 & 0.082 & 0.873 & 0.320 \\
\rowcolor{gray!10}
Ours & \mbox{H3 33B} & \textbf{22.03} & \textbf{69.25} & \textbf{0.028} & \textbf{33.77} & \textbf{0.972} & \textbf{3.176} & \textbf{0.923} & \textbf{0.031} & \textbf{0.887} & \textbf{0.393} \\
\bottomrule
\end{tabular}}
\caption{Teacher comparison on 150 paired cases under shared geometry and motion. EPE uses RAFT; V-Int. and P-Phys. are blind VLM visual-integrity and physical-plausibility pass rates (score $\geq8$). Comparisons are made only within the same backbone family; bold marks the best method in each family.}
\label{tab:teacher_main}
\end{table*}

\subsection{Asymmetric Physics-Guided Distribution Matching Distillation}
\label{sec:distillation}

We now use the physics-conditioned generator as a teacher, distilling its explicit physical guidance into an implicit generative prior through AP-DMD. We initialize the student from the H3 Turbo four-step model~\cite{lightx2v_h3_turbo2026} and apply DMD~\cite{yin2024onestepdiffusiondistributionmatching,yin2024improveddistributionmatchingdistillation}. The resulting student requires only the initial image and text, without simulation at inference.

Training uses three independent networks: the frozen physics-conditioned teacher $T_\phi$, the trainable student $G_\theta$, and a fake-score network $F_\psi$ that estimates the student's current output distribution. The fake-score network starts from the original H3 model with newly initialized LoRA parameters. The student and fake-score network use the same concatenated scene/action text, joint image--text features, self-attention conditioning, and first-frame latent prefix as the teacher. They receive only $c$ and have no PSS or SDS branch; the teacher additionally receives the paired conditions $h=(\mathbf P,\mathbf D)$.

\paragraph{Distribution-matching objective.}
We take the reverse KL divergence between the noised conditional distributions as the theoretical objective:
\begin{equation}
    \mathcal L_{\mathrm{DMD}}=
    \mathbb E_{(c,h),\sigma}\!\left[
        D_{\mathrm{KL}}\!\left(
            q_{\theta,\sigma}(\cdot\mid c)
            \,\|\,
            p_{\phi,\sigma}(\cdot\mid c,h)
        \right)\right],
    \label{eq:dmd_kl}
\end{equation}
where $q_{\theta,\sigma}$ and $p_{\phi,\sigma}$ are the student and teacher latent distributions after the same noising process. We construct updates from their denoising estimates without explicitly evaluating the KL value.

For each student update, $G_\theta$ first completes its four-step rollout, and we add independent noise to its output:
\begin{equation}
    \begin{aligned}
        y&=G_\theta(\epsilon_0;c),\\
        x_\sigma&=(1-\sigma)y+\sigma\epsilon,
        \quad \epsilon_0,\epsilon\overset{\mathrm{iid}}{\sim}\mathcal N(0,I).
    \end{aligned}
    \label{eq:dmd_noising}
\end{equation}
The teacher and fake-score network evaluate the same $x_\sigma$ at the same $\sigma$. Their velocity predictions give clean-latent estimates
\begin{equation}
    \begin{aligned}
        \hat z_T&=x_\sigma+\sigma v_\phi(x_\sigma,\sigma;c,h),\\
        \hat z_F&=x_\sigma+\sigma v_\psi(x_\sigma,\sigma;c).
    \end{aligned}
    \label{eq:dmd_clean_estimates}
\end{equation}
The teacher supplies a physics-conditioned estimate, while the fake-score network estimates denoising under the student's current distribution. The paired $h$ remains fixed and is not refitted to the student's generated trajectory.

Using the relation between denoising estimates and distribution scores, the KL gradient can be approximated as
\begin{equation}
    \nabla_\theta\mathcal L_{\mathrm{DMD}}\approx
    \mathbb E\!\left[
        \frac{(1-\sigma)^2}{\sigma^2}
        \left(\frac{\partial y}{\partial\theta}\right)^{\!\top}
        (\hat z_F-\hat z_T)
    \right].
    \label{eq:dmd_gradient}
\end{equation}
The two predictions are held fixed when constructing the student update, and only $G_\theta$ is optimized. Physical conditions therefore influence the student through the training signal without becoming student inputs.

\paragraph{Learning the student distribution.}
After updating the student, we hold it fixed and update the fake-score network using newly generated samples $y'$. With gradients stopped through $y'$, we form
\begin{equation}
    x'_\sigma=(1-\sigma)y'+\sigma\epsilon
\end{equation}
and optimize the flow-matching loss~\cite{lipman2023flowmatching}
\begin{equation}
    \mathcal L_F=\mathbb E\!\left[
        \operatorname{MSE}\!\left(
            v_\psi(x'_\sigma,\sigma;c),y'-\epsilon
        \right)\right].
    \label{eq:fake_loss}
\end{equation}
The student and fake-score network are updated alternately, with more frequent fake-score updates to track the changing student distribution. In practice, we use normalized updates and a memory-saving gradient approximation; implementation and optimization details are deferred to the appendix.

\renewcommand{\floatpagefraction}{0.65}
\renewcommand{\dblfloatpagefraction}{0.85}
\makeatletter
\setlength{\@dblfpsep}{8pt}
\setlength{\@dblfptop}{0pt}
\setlength{\@dblfpbot}{0pt plus 1fil}
\makeatother
\newsavebox{\teacherQualOriginal}
\newsavebox{\teacherQualEnlarged}
\begin{figure*}[!tp]
  \centering
  \sbox{\teacherQualOriginal}{\includegraphics[width=0.795\textwidth]{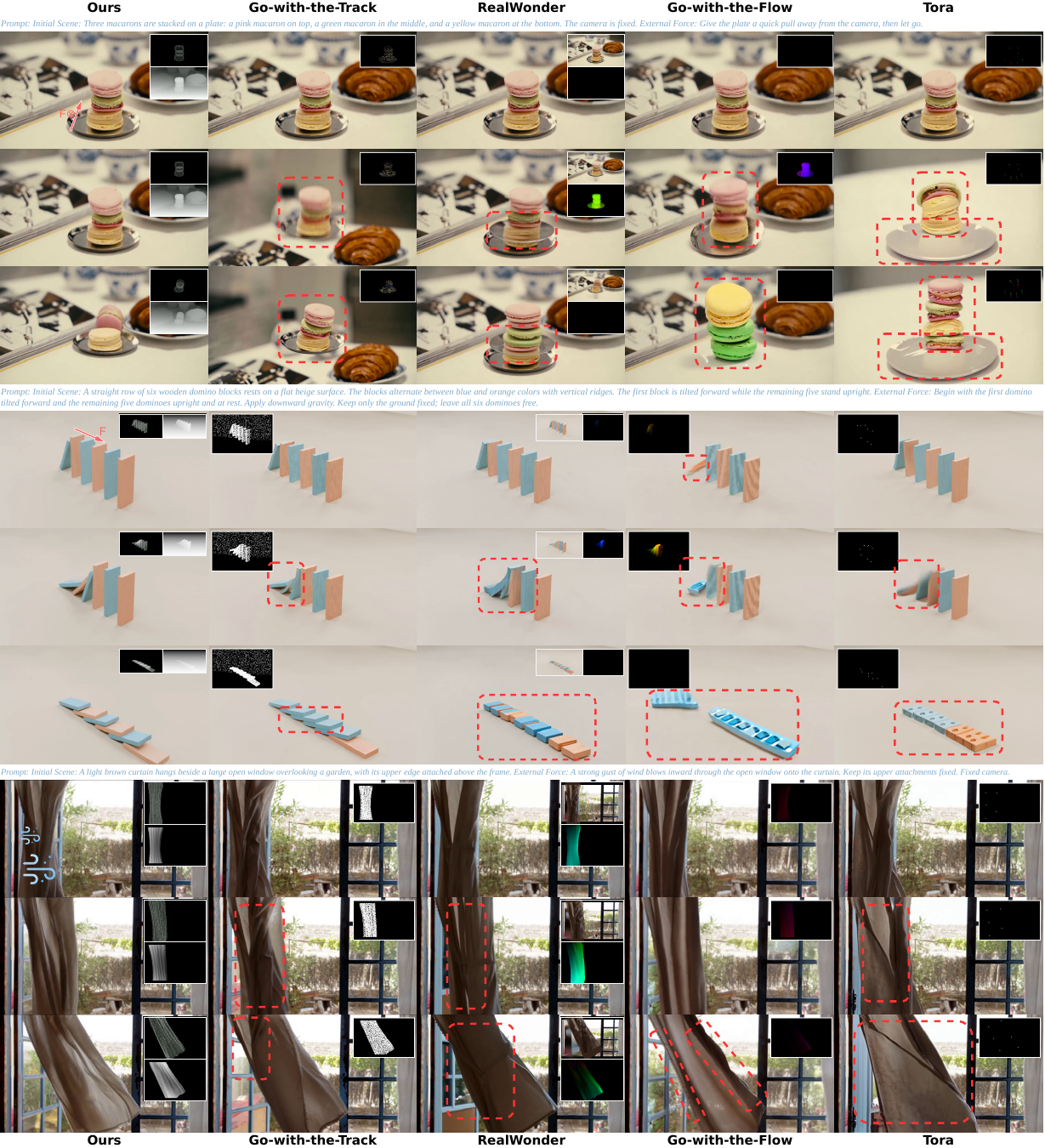}}
  \sbox{\teacherQualEnlarged}{\includegraphics[width=0.811\textwidth]{figures/teacher-qualitative-macaron-domino-curtain.pdf}}
  \usebox{\teacherQualEnlarged}
  \captionsetup{skip=2pt}
  \caption{Teacher comparison under shared geometry and motion. Insets show native controls; initial-frame arrows show prescribed forces. Red boxes highlight baseline failures.}
  \label{fig:teacher_qualitative}
  \vspace{\dimexpr\ht\teacherQualOriginal-\ht\teacherQualEnlarged+2pt\relax}
\end{figure*}

\begin{table*}[!tp]
\centering
\small
\setlength{\tabcolsep}{4pt}
\renewcommand{\arraystretch}{1.12}
\resizebox{\textwidth}{!}{%
\begin{tabular}{cccccccccc}
\toprule
\multirow{2}{*}{Method} & \multirow{2}{*}{Backbone} & \multirow{2}{*}{Steps} & \multicolumn{4}{c}{Temporal characteristics} & \multicolumn{2}{c}{Frame quality} & Physics-IQ Verified \\
\cmidrule(lr){4-7}\cmidrule(lr){8-9}\cmidrule(lr){10-10}
 & & & Subject $\uparrow$ & Backgr. $\uparrow$ & Smooth. $\uparrow$ & Dynamic & Aesthetic $\uparrow$ & Imaging $\uparrow$ & Score $\uparrow$ \\
\midrule
Wan2.2-TI2V & Wan2.2 5B & 50 & 90.89 & 94.11 & 99.21 & 47.95 & 41.49 & \underline{55.60} & 27.70 \\
\specialrule{0.8pt}{2pt}{2pt}
CogVideoX-I2V & CogVideoX 5B & 50 & 91.95 & 95.15 & 99.27 & 41.10 & 42.30 & 53.59 & 28.90 \\
\specialrule{0.8pt}{2pt}{2pt}
H3 Turbo & H3 33B & 4 & \underline{93.46} & \underline{95.59} & \underline{99.41} & 31.51 & \underline{43.90} & \textbf{60.06} & \underline{32.64} \\
\rowcolor{gray!10}
H3 Turbo + AP-DMD (Ours) & H3 33B & 4 & \textbf{95.42} & \textbf{96.81} & \textbf{99.63} & 12.33 & \textbf{44.74} & 50.06 & \textbf{40.00} \\
\bottomrule
\end{tabular}}
\caption{Student comparison on VBench and Physics-IQ Verified (best bold; second underlined; Dynamic unranked).}
\label{tab:student}
\end{table*}

\section{Experiments}
\label{sec:experiments}

\subsection{Experimental Setup}
\label{sec:experiment_setup}
\paragraph{Data and shared-scene protocol.}
We evaluate teachers on 150 paired cases, with 30 each covering 3D visibility and containment, multi-object contact propagation, localized force and actuation, deformable and rigid--soft contact, and basic rigid-body dynamics. Methods share the initial frame, English instruction, camera, motion, and seed; native controls are derived from the shared scene. Future RGB is reserved for scoring, and all generated cases are retained. Additional unpaired cases support qualitative inspection. Manifests and records are in the supplement.

\paragraph{Compared models and controls.}
We compare original H3, Tora~\cite{zhang2025toratrajectoryorienteddiffusiontransformer}, Go-with-the-Flow~\cite{burgert2025gowiththeflowmotioncontrollablevideodiffusion}, RealWonder~\cite{liu2026realwonderrealtimephysicalactionconditioned}, Go-with-the-Track~\cite{namekata2026gowiththetrackvideocompositingmotion}, and ParForce fine-tuned on the backbone families listed in the tables~\cite{wan22_release,wanfun13_release,wanfun22_release,videox_fun_release}. Our models receive a perspective particle state sequence and a surface-depth sequence; original H3 receives only the initial frame and text. Tora and Go-with-the-Track use projected tracks; Go-with-the-Flow uses flow-warped noise; RealWonder uses its released SDEdit~\cite{meng2022sdedit} and noise-conditioning pipeline. Sampling settings remain fixed across cases. Backbone-specific training and inference configurations are detailed in the supplement.

\paragraph{Common evaluation grid.}
We evaluate the same zero-to-three-second source-motion interval on a $384\times256$ full-view scoring grid. Frame indices follow physical motion time. Image metrics exclude the input frame and use eight equally spaced samples from $0.375$ to $3$ seconds. Per-video scores receive equal weight in each aggregate.

\subsection{Physics-Conditioned Teacher Comparison}
\label{sec:teacher_comparison}
\paragraph{Paired comparison.}
Table~\ref{tab:teacher_main} evaluates visual fidelity with FID/FVD~\cite{heusel2017fid,unterthiner2019fvd}, reconstruction with LPIPS/PSNR/SSIM~\cite{zhang2018lpips,wang2004ssim}, motion with RAFT EPE~\cite{teed2020raft}, and geometry with Canny F1 and relative-depth nRMSE. A blind VLM judge~\cite{qwen35_9b_release} assesses visual integrity and physical plausibility from the first frame, instruction, and generated video; we report pass rates at eight out of ten. Metric implementation details are provided in the supplementary material.

We rank methods within each backbone family and mark the best values in bold. ParForce leads all metrics on CogVideoX, Wan2.1 1.3B, and H3, and all except FID and relative-depth error on Wan2.2 14B. These matched-backbone comparisons control for generator scale and pretraining.
\begin{samepage}
\paragraph{Qualitative comparison.}
\textit{Macarons.} In Fig.~\ref{fig:teacher_qualitative}, Go-with-the-Track makes the macaron stack appear to pass through the rear obstacle, while the other baselines change object shape or appearance. Our joint-conditioned result better preserves geometry and spatial relations during the pull. \quad\textit{Dominoes.} Go-with-the-Track captures toppling but changes block colors; our result better maintains the contact sequence and alternating colors. \quad\textit{Curtain.} All methods capture inward motion, while our result shows more coherent folds and more natural changes in translucency.
\par\end{samepage}

\subsection{Simulator-Free Student Evaluation}
\label{sec:student_evaluation}
Our student uses only the initial frame and text. We initialize it from H3 Turbo~\cite{lightx2v_h3_turbo2026}, a released four-step LoRA, and compare the two at the same inference budget to assess the effect of AP-DMD.

Table~\ref{tab:student} reports six VBench~\cite{huang2023vbenchcomprehensivebenchmarksuite} dimensions on 146 equally weighted cases and Physics-IQ Verified~\cite{radsch2026physicsiqverified} on 198 views from 66 scenarios. We use the official prompts, preprocessing, mask generation, and Verified scoring code, reporting scores on a 0--100 scale. AP-DMD improves Physics-IQ Verified by 7.36 points over H3 Turbo at four-step inference.

\paragraph{Qualitative comparison.}
\textit{Stacked pebbles.} H3 Turbo duplicates the top pebble, while our student preserves a single top pebble throughout the motion. \quad\textit{Glass marbles.} H3 Turbo adds a ball, while our student better preserves object count and glass appearance during contact.

\begin{figure}[!htbp]
  \centering
  \includegraphics[width=\columnwidth]{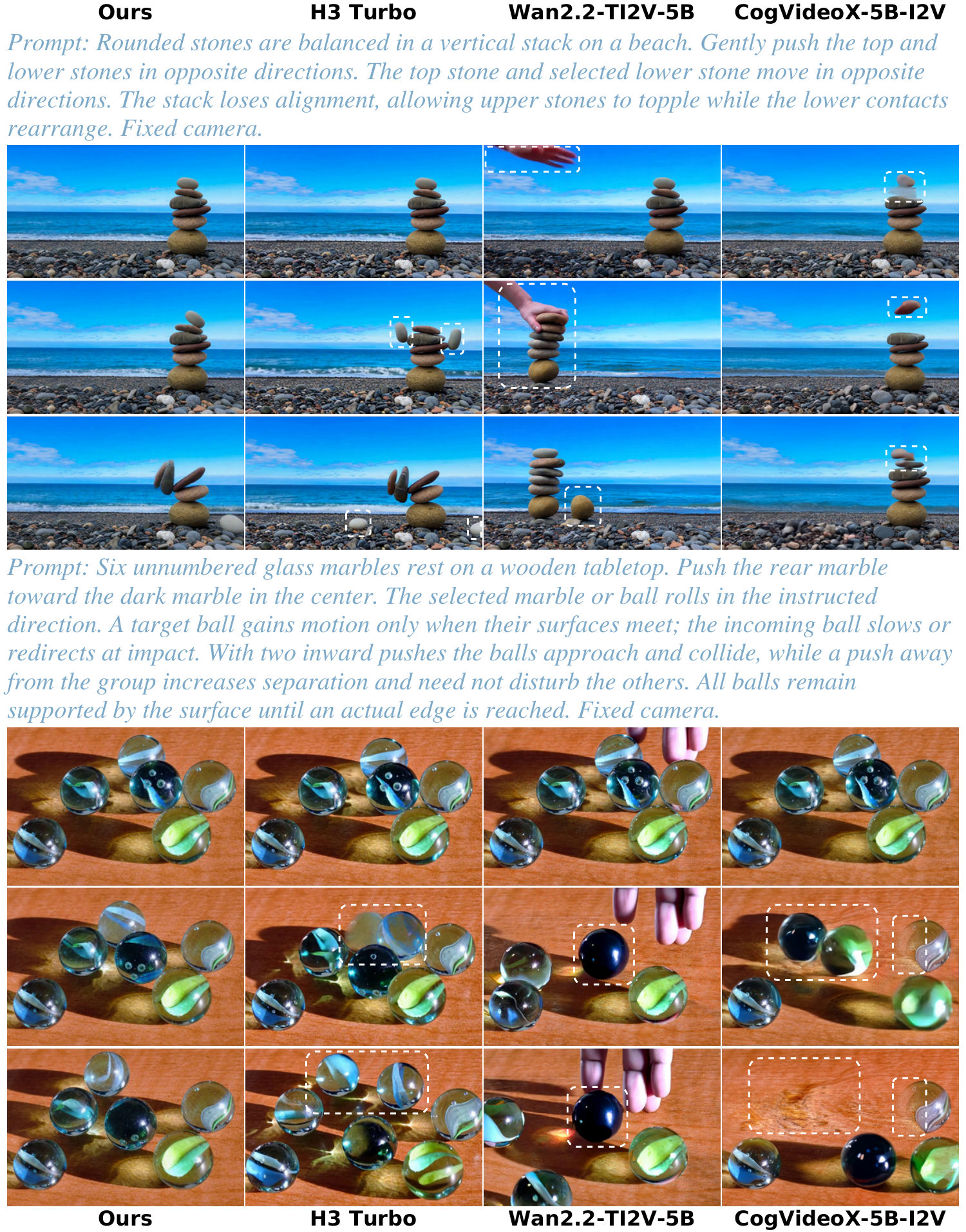}
  \caption{Simulator-free student comparison on two cases. (Boxes highlight implausible results from the other baselines.)}
  \label{fig:student_qualitative}
\end{figure}

\subsection{Control-Modality Ablation}
\label{sec:control_modality_ablation}
\begin{figure}[!htbp]
  \centering
  \includegraphics[width=\columnwidth]{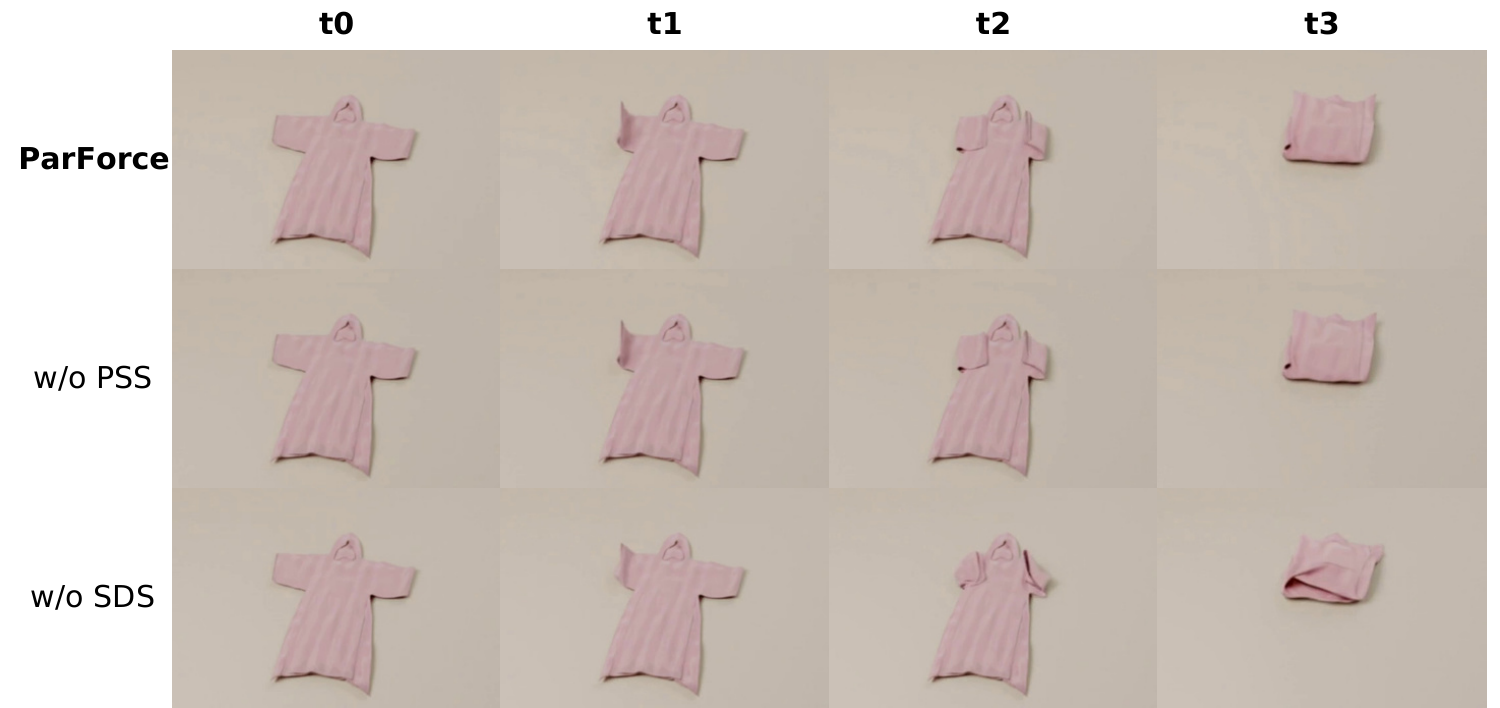}
  \caption{Control-modality ablation on garment folding. Rows: ParForce, w/o PSS, and w/o SDS. Columns show synchronized times.}
  \label{fig:particle-depth-ablation}
\end{figure}

We remove PSS or SDS and train each variant with the same data and training budget. All three checkpoints use matched inference settings and a common 49-frame alignment on 30 test cases. Full ParForce has the best means across all four metrics (Table~\ref{tab:conditioning_ablation}), with small margins in subject consistency and imaging quality. PSS supplies persistent subject-motion correspondences, while SDS supplies continuous geometry and visibility boundaries. Motion IoU uses a source-RGB RAFT reference; subject consistency and imaging quality use VBench I2V subject~\cite{huang2024vbenchplusplus} and MUSIQ~\cite{ke2021musiq}. Figure~\ref{fig:particle-depth-ablation} shows the effect on hooded-garment~\cite{korosteleva2021garments} folding.
\begin{table}[H]
\centering
\small
\renewcommand{\arraystretch}{1.12}
\setlength{\tabcolsep}{4pt}
\resizebox{\columnwidth}{!}{%
\begin{tabular}{@{}ccccc@{}}
\toprule
Variant & EPE $\downarrow$ (px) & Motion IoU $\uparrow$ & Subject $\uparrow$ & Imaging $\uparrow$ \\
\midrule
\textbf{ParForce} & \textbf{9.05} & \textbf{43.62} & \textbf{96.35} & \textbf{57.20} \\
w/o PSS (SDS only) & 12.96 & 43.08 & 96.33 & 57.13 \\
w/o SDS (PSS only) & 14.06 & 43.20 & 95.96 & 57.16 \\
\bottomrule
\end{tabular}%
}
\par\smallskip
\caption{Control-modality ablation. PSS: perspective particle state sequence; SDS: surface-depth sequence.}
\label{tab:conditioning_ablation}
\end{table}

\subsection{Effect of Real-Video Mixing}
\label{sec:real_video_mixing}
\begin{figure}[!htbp]
  \centering
  \includegraphics[width=\columnwidth]{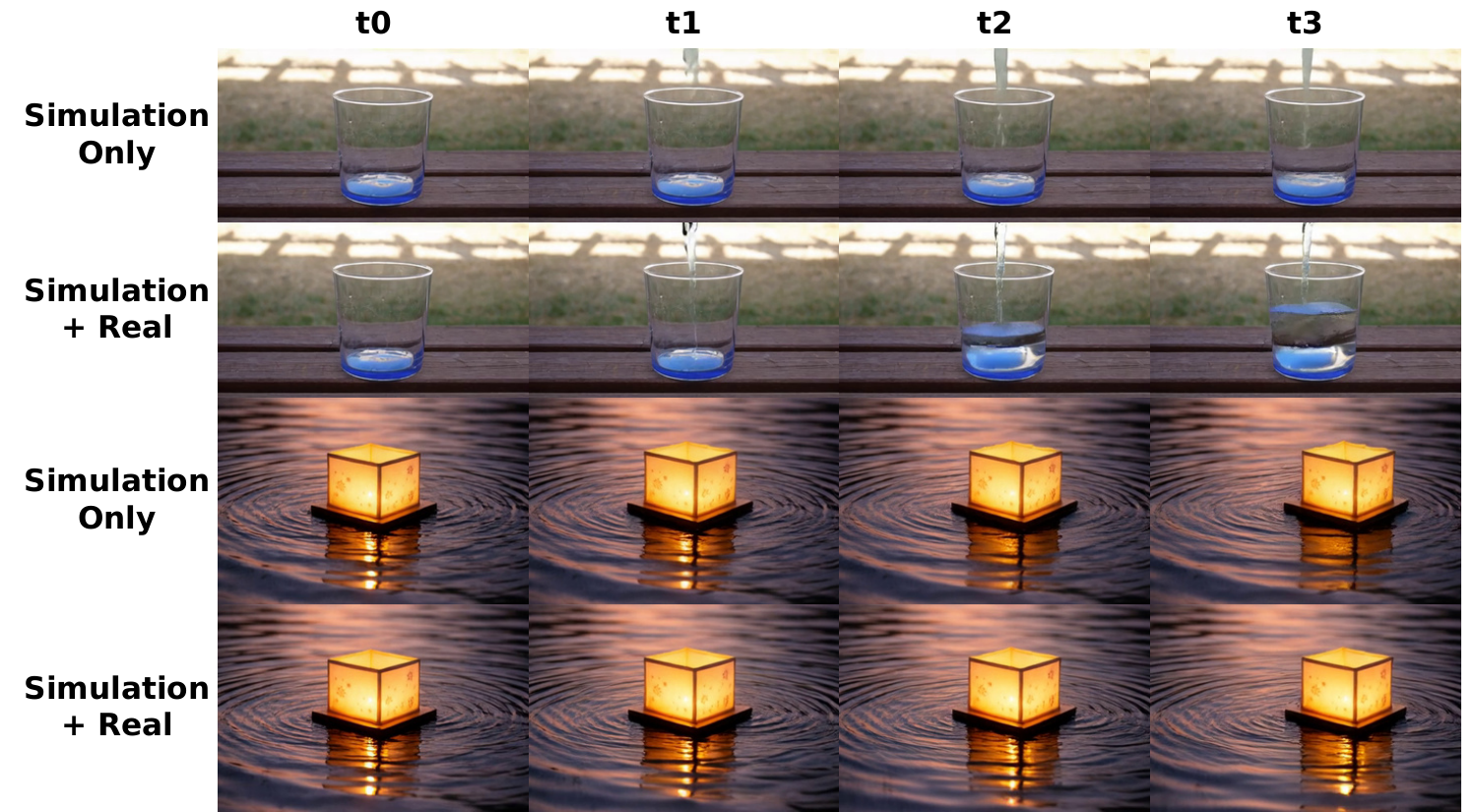}
  \caption{Effect of real-video mixing.}
  \label{fig:real_video_mixing}
\end{figure}

We compare simulation-only and 1:1 mixed training with depth-conditioned real videos, using Wan2.1-Fun 1.3B with matched training budgets and inference inputs. Figure~\ref{fig:real_video_mixing} illustrates the differences: \textit{Water pouring.} Mixing produces a more natural water column and clearer water-level rise. \textit{Floating lantern.} Simulation-only ripples remain almost static, while mixed training produces more natural ripple evolution. These results support real-video mixing as a means of narrowing the sim-to-real gap.

\subsection{Robustness to Particle Conditioning}
\label{sec:particle_robustness}
We vary particle density (64--1024 points) and geometric shape, including simplified abstractions, while fixing the initial frame, text, depth, camera, and physical motion. Generation metrics vary little across particle counts; the full results are provided in the supplementary material. Figure~\ref{fig:particle_density} compares the count endpoints and geometric abstractions, illustrating tolerance to changes in the particle representation.
\begin{figure}[!htbp]
  \centering
  \includegraphics[width=\columnwidth]{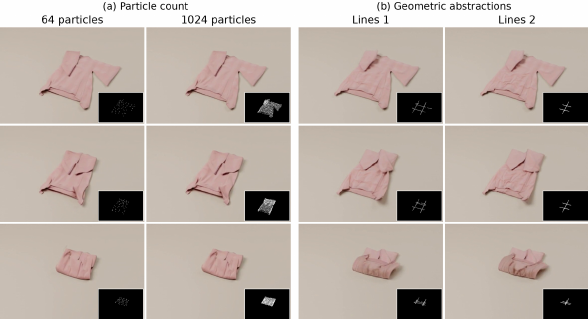}
  \caption{Particle-count variation (left) and geometric abstractions (right).}
  \label{fig:particle_density}
\end{figure}

\FloatBarrier
\section{Conclusion}
\label{sec:conclusion}
ParForce guides video generation with PSS and SDS. Mixed simulated RGB--particle--depth and real RGB--depth training narrows the sim-to-real appearance gap without real-world particle annotations. Across backbones, it improves visual, reconstruction, and motion fidelity; physics-guided distillation transfers this guidance to four-step generation from an initial image and text.

\clearpage
\paragraph*{Acknowledgments.}
OpenAI Codex assisted with code implementation and editing of author-produced material.

{\small
\linespread{0.92}\selectfont
\setlength{\bibsep}{0pt}
\bibliographystyle{ieeenat_fullname}
\bibliography{references}

\begin{thebibliography}{48}
\providecommand{\natexlab}[1]{#1}
\providecommand{\url}[1]{\texttt{#1}}
\expandafter\ifx\csname urlstyle\endcsname\relax
  \providecommand{\doi}[1]{doi: #1}\else
  \providecommand{\doi}{doi: \begingroup \urlstyle{rm}\Url}\fi

\bibitem[{Alibaba PAI}(2025{\natexlab{a}})]{wanfun13_release}
{Alibaba PAI}.
\newblock Wan2.1-fun-v1.1-1.3b-control.
\newblock
  \url{https://huggingface.co/alibaba-pai/Wan2.1-Fun-V1.1-1.3B-Control},
  2025{\natexlab{a}}.
\newblock Accessed November 16, 2026.

\bibitem[{Alibaba PAI}(2025{\natexlab{b}})]{wanfun22_release}
{Alibaba PAI}.
\newblock Wan2.2-fun-a14b-control.
\newblock \url{https://huggingface.co/alibaba-pai/Wan2.2-Fun-A14B-Control},
  2025{\natexlab{b}}.
\newblock Accessed November 16, 2026.

\bibitem[{Alibaba PAI}(2026)]{h3_fun_union2_release}
{Alibaba PAI}.
\newblock {MiniMax-H3-Fun-Controlnet-Union-2.0}.
\newblock
  \url{https://huggingface.co/alibaba-pai/MiniMax-H3-Fun-Controlnet-Union-2.0},
  2026.

\bibitem[Burgert et~al.(2025)Burgert, Xu, Xian, Pilarski, Clausen, He, Ma,
  Deng, Li, Mousavi, Ryoo, Debevec, and
  Yu]{burgert2025gowiththeflowmotioncontrollablevideodiffusion}
Ryan Burgert, Yuancheng Xu, Wenqi Xian, Oliver Pilarski, Pascal Clausen,
  Mingming He, Li Ma, Yitong Deng, Lingxiao Li, Mohsen Mousavi, Michael Ryoo,
  Paul Debevec, and Ning Yu.
\newblock {Go-with-the-Flow}: Motion-controllable video diffusion models using
  real-time warped noise.
\newblock In \emph{Proceedings of the IEEE/CVF Conference on Computer Vision
  and Pattern Recognition}, 2025.

\bibitem[Chu et~al.(2025)Chu, He, Chen, Zhang, Xu, Xia, Wang, Yi, Liu, Zhao,
  Liu, Zhang, and Yang]{chu2025wanmovemotioncontrollablevideogeneration}
Ruihang Chu, Yefei He, Zhekai Chen, Shiwei Zhang, Xiaogang Xu, Bin Xia,
  Dingdong Wang, Hongwei Yi, Xihui Liu, Hengshuang Zhao, Yu Liu, Yingya Zhang,
  and Yujiu Yang.
\newblock {Wan-Move}: Motion-controllable video generation via latent
  trajectory guidance.
\newblock arXiv preprint arXiv:2512.08765, 2025.

\bibitem[Feng et~al.(2026)Feng, Karkus, Igl, Berner, Chen, Tan, Alahi,
  Ivanovic, and Pavone]{feng2026parametrictrajectorydistillationfewstep}
Lan Feng, Peter Karkus, Maximilian Igl, Julius Berner, Yuxiao Chen, Shuhan Tan,
  Alexandre Alahi, Boris Ivanovic, and Marco Pavone.
\newblock Parametric trajectory distillation for few-step video generation.
\newblock arXiv preprint arXiv:2610.11498, 2026.

\bibitem[Gu et~al.(2025)Gu, Yan, Lu, Li, Dou, Si, Dong, Liu, Lin, Liu, Wang,
  and Liu]{gu2025diffusionshader3dawarevideo}
Zekai Gu, Rui Yan, Jiahao Lu, Peng Li, Zhiyang Dou, Chenyang Si, Zhen Dong,
  Qifeng Liu, Cheng Lin, Ziwei Liu, Wenping Wang, and Yuan Liu.
\newblock Diffusion as shader: {3D}-aware video diffusion for versatile video
  generation control.
\newblock In \emph{ACM SIGGRAPH Conference Papers}, 2025.

\bibitem[Gupta et~al.(2015)Gupta, Hoffman, and
  Malik]{gupta2015crossmodaldistillationsupervision}
Saurabh Gupta, Judy Hoffman, and Jitendra Malik.
\newblock Cross modal distillation for supervision transfer.
\newblock arXiv preprint arXiv:1507.00448, 2015.

\bibitem[Heusel et~al.(2017)Heusel, Ramsauer, Unterthiner, Nessler, and
  Hochreiter]{heusel2017fid}
Martin Heusel, Hubert Ramsauer, Thomas Unterthiner, Bernhard Nessler, and Sepp
  Hochreiter.
\newblock {GANs} trained by a two time-scale update rule converge to a local
  nash equilibrium.
\newblock In \emph{Advances in Neural Information Processing Systems}, 2017.

\bibitem[Hu et~al.(2022)Hu, Shen, Wallis, Allen-Zhu, Li, Wang, Wang, and
  Chen]{hu2022lora}
Edward~J. Hu, Yelong Shen, Phillip Wallis, Zeyuan Allen-Zhu, Yuanzhi Li, Shean
  Wang, Lu Wang, and Weizhu Chen.
\newblock {LoRA}: Low-rank adaptation of large language models.
\newblock In \emph{International Conference on Learning Representations}, 2022.

\bibitem[Huang et~al.(2024{\natexlab{a}})Huang, He, Yu, Zhang, Si, Jiang,
  Zhang, Wu, Jin, Chanpaisit, Wang, Chen, Wang, Lin, Qiao, and
  Liu]{huang2023vbenchcomprehensivebenchmarksuite}
Ziqi Huang, Yinan He, Jiashuo Yu, Fan Zhang, Chenyang Si, Yuming Jiang, Yuanhan
  Zhang, Tianxing Wu, Qingyang Jin, Nattapol Chanpaisit, Yaohui Wang, Xinyuan
  Chen, Limin Wang, Dahua Lin, Yu Qiao, and Ziwei Liu.
\newblock {VBench}: Comprehensive benchmark suite for video generative models.
\newblock In \emph{Proceedings of the IEEE/CVF Conference on Computer Vision
  and Pattern Recognition}, 2024{\natexlab{a}}.

\bibitem[Huang et~al.(2024{\natexlab{b}})Huang, Zhang, Xu, He, Yu, Dong, Ma,
  Chanpaisit, Si, Jiang, Wang, Chen, Chen, Wang, Lin, Qiao, and
  Liu]{huang2024vbenchplusplus}
Ziqi Huang, Fan Zhang, Xiaojie Xu, Yinan He, Jiashuo Yu, Ziyue Dong, Qianli Ma,
  Nattapol Chanpaisit, Chenyang Si, Yuming Jiang, Yaohui Wang, Xinyuan Chen,
  Ying-Cong Chen, Limin Wang, Dahua Lin, Yu Qiao, and Ziwei Liu.
\newblock {VBench++}: Comprehensive and versatile benchmark suite for video
  generative models.
\newblock arXiv preprint arXiv:2411.13503, 2024{\natexlab{b}}.

\bibitem[Ke et~al.(2021)Ke, Wang, Wang, Milanfar, and Yang]{ke2021musiq}
Junjie Ke, Qifei Wang, Yilin Wang, Peyman Milanfar, and Feng Yang.
\newblock {MUSIQ}: Multi-scale image quality transformer.
\newblock In \emph{Proceedings of the IEEE/CVF International Conference on
  Computer Vision}, 2021.

\bibitem[Kong et~al.(2025)Kong, Tian, Zhang, Min, Dai, Zhou, Xiong, Li, Wu,
  Zhang, Wu, Lin, Yuan, Long, Wang, Wang, Li, Huang, Yang, Tan, Wang, Song,
  Bai, Wu, Xue, Wang, Wang, Liu, Li, Li, Wang, Yu, Deng, Li, Chen, Cui, Peng,
  Yu, He, Xu, Zhou, Xu, Tao, Lu, Liu, Zhou, Wang, Yang, Wang, Liu, Jiang, and
  Zhong]{kong2025hunyuanvideosystematicframeworklarge}
Weijie Kong, Qi Tian, Zijian Zhang, Rox Min, Zuozhuo Dai, Jin Zhou, Jiangfeng
  Xiong, Xin Li, Bo Wu, Jianwei Zhang, Kathrina Wu, Qin Lin, Junkun Yuan,
  Yanxin Long, Aladdin Wang, Andong Wang, Changlin Li, Duojun Huang, Fang Yang,
  Hao Tan, Hongmei Wang, Jacob Song, Jiawang Bai, Jianbing Wu, Jinbao Xue, Joey
  Wang, Kai Wang, Mengyang Liu, Pengyu Li, Shuai Li, Weiyan Wang, Wenqing Yu,
  Xinchi Deng, Yang Li, Yi Chen, Yutao Cui, Yuanbo Peng, Zhentao Yu, Zhiyu He,
  Zhiyong Xu, Zixiang Zhou, Zunnan Xu, Yangyu Tao, Qinglin Lu, Songtao Liu, Dax
  Zhou, Hongfa Wang, Yong Yang, Di Wang, Yuhong Liu, Jie Jiang, and Caesar
  Zhong.
\newblock {HunyuanVideo}: A systematic framework for large video generative
  models.
\newblock arXiv preprint arXiv:2412.03603, 2025.

\bibitem[Korosteleva and Lee(2021)]{korosteleva2021garments}
Maria Korosteleva and Sung-Hee Lee.
\newblock Generating datasets of {3D} garments with sewing patterns.
\newblock In \emph{Advances in Neural Information Processing Systems Datasets
  and Benchmarks Track}, 2021.

\bibitem[Li et~al.(2025)Li, Yu, Liu, Yang, Herrmann, Wetzstein, and
  Wu]{li2025wonderplaydynamic3dscene}
Zizhang Li, Hong-Xing Yu, Wei Liu, Yin Yang, Charles Herrmann, Gordon
  Wetzstein, and Jiajun Wu.
\newblock {WonderPlay}: Dynamic {3D} scene generation from a single image and
  actions.
\newblock In \emph{Proceedings of the IEEE/CVF International Conference on
  Computer Vision}, 2025.

\bibitem[Lipman et~al.(2023)Lipman, Chen, Ben-Hamu, Nickel, and
  Le]{lipman2023flowmatching}
Yaron Lipman, Ricky T.~Q. Chen, Heli Ben-Hamu, Maximilian Nickel, and Matt Le.
\newblock Flow matching for generative modeling.
\newblock In \emph{International Conference on Learning Representations}, 2023.

\bibitem[Liu et~al.(2024)Liu, Ren, Gupta, and
  Wang]{liu2024physgenrigidbodyphysicsgroundedimagetovideo}
Shaowei Liu, Zhongzheng Ren, Saurabh Gupta, and Shenlong Wang.
\newblock {PhysGen}: Rigid-body physics-grounded image-to-video generation.
\newblock In \emph{European Conference on Computer Vision}, 2024.

\bibitem[Liu et~al.(2026)Liu, Chen, Li, Wang, Yu, and
  Wu]{liu2026realwonderrealtimephysicalactionconditioned}
Wei Liu, Ziyu Chen, Zizhang Li, Yue Wang, Hong-Xing Yu, and Jiajun Wu.
\newblock {RealWonder}: Real-time physical action-conditioned video generation.
\newblock arXiv preprint arXiv:2603.05449, 2026.

\bibitem[Lopez-Paz et~al.(2016)Lopez-Paz, Bottou, Schölkopf, and
  Vapnik]{lopezpaz2016unifyingdistillationprivilegedinformation}
David Lopez-Paz, Léon Bottou, Bernhard Schölkopf, and Vladimir Vapnik.
\newblock Unifying distillation and privileged information.
\newblock In \emph{International Conference on Learning Representations}, 2016.

\bibitem[Meng et~al.(2022)Meng, He, Song, Song, Wu, Zhu, and
  Ermon]{meng2022sdedit}
Chenlin Meng, Yutong He, Yang Song, Jiaming Song, Jiajun Wu, Jun-Yan Zhu, and
  Stefano Ermon.
\newblock {SDEdit}: Guided image synthesis and editing with stochastic
  differential equations.
\newblock In \emph{International Conference on Learning Representations}, 2022.

\bibitem[{MiniMax}(2026)]{minimax_h3_release}
{MiniMax}.
\newblock {MiniMax H3}.
\newblock \url{https://huggingface.co/MiniMaxAI/MiniMax-H3}, 2026.
\newblock Accessed November 16, 2026.

\bibitem[{ModelTC}(2026)]{lightx2v_h3_turbo2026}
{ModelTC}.
\newblock {MiniMax-H3 Turbo}.
\newblock \url{https://huggingface.co/lightx2v/Minimax-h3-Turbo}, 2026.
\newblock Accessed 2026-11-09.

\bibitem[Motamed et~al.(2026)Motamed, Culp, Swersky, Jaini, and
  Geirhos]{motamed2026physicsiq}
Saman Motamed, Laura Culp, Kevin Swersky, Priyank Jaini, and Robert Geirhos.
\newblock Do generative video models understand physical principles?
\newblock In \emph{Proceedings of the IEEE/CVF Winter Conference on
  Applications of Computer Vision}, pages 948--958, 2026.

\bibitem[Namekata et~al.(2026)Namekata, Kant, Liu, Burgert, Xu, Lin, Steven,
  Philip, Ma, Vedaldi, Debevec, and
  Yu]{namekata2026gowiththetrackvideocompositingmotion}
Koichi Namekata, Yash Kant, Zhizheng Liu, Ryan~D Burgert, Yuancheng Xu,
  Kuan~Heng Lin, Emmett Steven, Julien Philip, Li Ma, Andrea Vedaldi, Paul
  Debevec, and Ning Yu.
\newblock {Go-with-the-Track}: Video compositing and motion control with point
  tracking.
\newblock In \emph{ACM SIGGRAPH Conference Papers}, 2026.

\bibitem[{Qwen Team}(2026{\natexlab{a}})]{qwen35_9b_release}
{Qwen Team}.
\newblock {Qwen3.5-9B}.
\newblock \url{https://huggingface.co/Qwen/Qwen3.5-9B}, 2026{\natexlab{a}}.
\newblock Accessed November 16, 2026.

\bibitem[{Qwen Team}(2026{\natexlab{b}})]{qwen35_release}
{Qwen Team}.
\newblock {Qwen3.5}: Accelerating productivity with native multimodal agents.
\newblock \url{https://qwen.ai/blog?id=qwen3.5}, 2026{\natexlab{b}}.

\bibitem[Ravi et~al.(2025)Ravi, Gabeur, Hu, Hu, Ryali, Ma, Khedr, Rädle,
  Rolland, Gustafson, Mintun, Pan, Alwala, Carion, Wu, Girshick, Dollár, and
  Feichtenhofer]{ravi2024sam2}
Nikhila Ravi, Valentin Gabeur, Yuan-Ting Hu, Ronghang Hu, Chaitanya Ryali,
  Tengyu Ma, Haitham Khedr, Roman Rädle, Chloe Rolland, Laura Gustafson, Eric
  Mintun, Junting Pan, Kalyan~Vasudev Alwala, Nicolas Carion, Chao-Yuan Wu,
  Ross Girshick, Piotr Dollár, and Christoph Feichtenhofer.
\newblock {SAM 2}: Segment anything in images and videos.
\newblock In \emph{International Conference on Learning Representations}, 2025.

\bibitem[Rädsch et~al.(2026)Rädsch, Asano, Kuehne, Bauer, Jaini, Geirhos, and
  Lüth]{radsch2026physicsiqverified}
Tim Rädsch, Yuki~M. Asano, Hilde Kuehne, Stefan Bauer, Priyank Jaini, Robert
  Geirhos, and Carsten~T. Lüth.
\newblock {Physics-IQ Verified}.
\newblock \emph{arXiv preprint arXiv:2606.18943}, 2026.

\bibitem[Salimans and
  Ho(2022)]{salimans2022progressivedistillationfastsampling}
Tim Salimans and Jonathan Ho.
\newblock Progressive distillation for fast sampling of diffusion models.
\newblock In \emph{International Conference on Learning Representations}, 2022.

\bibitem[Song et~al.(2023)Song, Dhariwal, Chen, and
  Sutskever]{song2023consistencymodels}
Yang Song, Prafulla Dhariwal, Mark Chen, and Ilya Sutskever.
\newblock Consistency models.
\newblock In \emph{International Conference on Machine Learning}, 2023.

\bibitem[Tan et~al.(2024)Tan, Jiang, Li, Zong, Xie, Yang, and
  Jiang]{tan2024physmotionphysicsgroundeddynamicssingle}
Xiyang Tan, Ying Jiang, Xuan Li, Zeshun Zong, Tianyi Xie, Yin Yang, and
  Chenfanfu Jiang.
\newblock {PhysMotion}: Physics-grounded dynamics from a single image.
\newblock arXiv preprint arXiv:2411.17189, 2024.

\bibitem[Teed and Deng(2020)]{teed2020raft}
Zachary Teed and Jia Deng.
\newblock {RAFT}: Recurrent all-pairs field transforms for optical flow.
\newblock In \emph{European Conference on Computer Vision}, 2020.

\bibitem[Todorov et~al.(2012)Todorov, Erez, and Tassa]{todorov2012mujoco}
Emanuel Todorov, Tom Erez, and Yuval Tassa.
\newblock {MuJoCo}: A physics engine for model-based control.
\newblock In \emph{IEEE/RSJ International Conference on Intelligent Robots and
  Systems}, pages 5026--5033, 2012.

\bibitem[Unterthiner et~al.(2019)Unterthiner, van Steenkiste, Kurach, Marinier,
  Michalski, and Gelly]{unterthiner2019fvd}
Thomas Unterthiner, Sjoerd van Steenkiste, Karol Kurach, Raphael Marinier,
  Marcin Michalski, and Sylvain Gelly.
\newblock Towards accurate generative models of video: A new metric \&
  challenges.
\newblock arXiv preprint arXiv:1812.01717, 2019.

\bibitem[{VideoX-Fun Contributors}(2026)]{videox_fun_release}
{VideoX-Fun Contributors}.
\newblock {VideoX-Fun}.
\newblock \url{https://github.com/aigc-apps/VideoX-Fun}, 2026.
\newblock Accessed November 16, 2026.

\bibitem[Wan et~al.(2025)Wan, Wang, Ai, Wen, Mao, Xie, Chen, Yu, Zhao, Yang,
  Zeng, Wang, Zhang, Zhou, Wang, Chen, Zhu, Zhao, Yan, Huang, Feng, Zhang, Li,
  Wu, Chu, Feng, Zhang, Sun, Fang, Wang, Gui, Weng, Shen, Lin, Wang, Wang,
  Zhou, Wang, Shen, Yu, Shi, Huang, Xu, Kou, Lv, Li, Liu, Wang, Zhang, Huang,
  Li, Wu, Liu, Pan, Zheng, Hong, Shi, Feng, Jiang, Han, Wu, and
  Liu]{wan2025wanopenadvancedlargescale}
Team Wan, Ang Wang, Baole Ai, Bin Wen, Chaojie Mao, Chen-Wei Xie, Di Chen,
  Feiwu Yu, Haiming Zhao, Jianxiao Yang, Jianyuan Zeng, Jiayu Wang, Jingfeng
  Zhang, Jingren Zhou, Jinkai Wang, Jixuan Chen, Kai Zhu, Kang Zhao, Keyu Yan,
  Lianghua Huang, Mengyang Feng, Ningyi Zhang, Pandeng Li, Pingyu Wu, Ruihang
  Chu, Ruili Feng, Shiwei Zhang, Siyang Sun, Tao Fang, Tianxing Wang, Tianyi
  Gui, Tingyu Weng, Tong Shen, Wei Lin, Wei Wang, Wei Wang, Wenmeng Zhou, Wente
  Wang, Wenting Shen, Wenyuan Yu, Xianzhong Shi, Xiaoming Huang, Xin Xu, Yan
  Kou, Yangyu Lv, Yifei Li, Yijing Liu, Yiming Wang, Yingya Zhang, Yitong
  Huang, Yong Li, You Wu, Yu Liu, Yulin Pan, Yun Zheng, Yuntao Hong, Yupeng
  Shi, Yutong Feng, Zeyinzi Jiang, Zhen Han, Zhi-Fan Wu, and Ziyu Liu.
\newblock Wan: Open and advanced large-scale video generative models.
\newblock arXiv preprint arXiv:2503.20314, 2025.

\bibitem[{Wan Team}(2025)]{wan22_release}
{Wan Team}.
\newblock {Wan2.2}.
\newblock \url{https://github.com/Wan-Video/Wan2.2}, 2025.
\newblock Accessed November 16, 2026.

\bibitem[Wang et~al.(2025)Wang, Xu, Dai, Xiang, Deng, Tong, and
  Yang]{wang2025moge}
Ruicheng Wang, Sicheng Xu, Cassie Dai, Jianfeng Xiang, Yu Deng, Xin Tong, and
  Jiaolong Yang.
\newblock {MoGe}: Unlocking accurate monocular geometry estimation for
  open-domain images with optimal training supervision.
\newblock In \emph{Proceedings of the IEEE/CVF Conference on Computer Vision
  and Pattern Recognition}, 2025.

\bibitem[Wang et~al.(2004)Wang, Bovik, Sheikh, and Simoncelli]{wang2004ssim}
Zhou Wang, Alan~Conrad Bovik, Hamid~Rahim Sheikh, and Eero~P. Simoncelli.
\newblock Image quality assessment: From error visibility to structural
  similarity.
\newblock \emph{IEEE Transactions on Image Processing}, 13\penalty0
  (4):\penalty0 600--612, 2004.

\bibitem[Yang et~al.(2024)Yang, Kang, Huang, Zhao, Xu, Feng, and
  Zhao]{yang2024depthanythingv2}
Lihe Yang, Bingyi Kang, Zilong Huang, Zhen Zhao, Xiaogang Xu, Jiashi Feng, and
  Hengshuang Zhao.
\newblock Depth anything v2.
\newblock In \emph{Advances in Neural Information Processing Systems}, 2024.

\bibitem[Yang et~al.(2025)Yang, Teng, Zheng, Ding, Huang, Xu, Yang, Hong,
  Zhang, Feng, Yin, Zhang, Wang, Cheng, Xu, Gu, Dong, and
  Tang]{yang2025cogvideoxtexttovideodiffusionmodels}
Zhuoyi Yang, Jiayan Teng, Wendi Zheng, Ming Ding, Shiyu Huang, Jiazheng Xu,
  Yuanming Yang, Wenyi Hong, Xiaohan Zhang, Guanyu Feng, Da Yin, Yuxuan Zhang,
  Weihan Wang, Yean Cheng, Bin Xu, Xiaotao Gu, Yuxiao Dong, and Jie Tang.
\newblock {CogVideoX}: Text-to-video diffusion models with an expert
  transformer.
\newblock arXiv preprint arXiv:2408.06072, 2025.

\bibitem[Yin et~al.(2024{\natexlab{a}})Yin, Gharbi, Park, Zhang, Shechtman,
  Durand, and Freeman]{yin2024improveddistributionmatchingdistillation}
Tianwei Yin, Michaël Gharbi, Taesung Park, Richard Zhang, Eli Shechtman, Fredo
  Durand, and William~T. Freeman.
\newblock Improved distribution matching distillation for fast image synthesis.
\newblock In \emph{Advances in Neural Information Processing Systems},
  2024{\natexlab{a}}.

\bibitem[Yin et~al.(2024{\natexlab{b}})Yin, Gharbi, Zhang, Shechtman, Durand,
  Freeman, and Park]{yin2024onestepdiffusiondistributionmatching}
Tianwei Yin, Michaël Gharbi, Richard Zhang, Eli Shechtman, Fredo Durand,
  William~T. Freeman, and Taesung Park.
\newblock One-step diffusion with distribution matching distillation.
\newblock In \emph{Proceedings of the IEEE/CVF Conference on Computer Vision
  and Pattern Recognition}, 2024{\natexlab{b}}.

\bibitem[Zhang et~al.(2023)Zhang, Rao, and Agrawala]{zhang2023controlnet}
Lvmin Zhang, Anyi Rao, and Maneesh Agrawala.
\newblock Adding conditional control to text-to-image diffusion models.
\newblock In \emph{Proceedings of the IEEE/CVF International Conference on
  Computer Vision}, 2023.

\bibitem[Zhang et~al.(2018)Zhang, Isola, Efros, Shechtman, and
  Wang]{zhang2018lpips}
Richard Zhang, Phillip Isola, Alexei~A. Efros, Eli Shechtman, and Oliver Wang.
\newblock The unreasonable effectiveness of deep features as a perceptual
  metric.
\newblock In \emph{Proceedings of the IEEE Conference on Computer Vision and
  Pattern Recognition}, 2018.

\bibitem[Zhang et~al.(2025)Zhang, Liao, Li, Dai, Qiu, Zhu, Qin, and
  Wang]{zhang2025toratrajectoryorienteddiffusiontransformer}
Zhenghao Zhang, Junchao Liao, Menghao Li, Zuozhuo Dai, Bingxue Qiu, Siyu Zhu,
  Long Qin, and Weizhi Wang.
\newblock Tora: Trajectory-oriented diffusion transformer for video generation.
\newblock In \emph{Proceedings of the IEEE/CVF Conference on Computer Vision
  and Pattern Recognition}, 2025.

\bibitem[Zhou et~al.(2024)Zhou, Zheng, Wang, Yin, and
  Huang]{zhou2024scoreidentitydistillationexponentially}
Mingyuan Zhou, Huangjie Zheng, Zhendong Wang, Mingzhang Yin, and Hai Huang.
\newblock Score identity distillation: Exponentially fast distillation of
  pretrained diffusion models for one-step generation.
\newblock In \emph{International Conference on Machine Learning}, 2024.

\end{thebibliography}
}

\clearpage
\flushbottom
\appendix
\twocolumn[{%
\pdfbookmark[0]{Appendix}{parforce-appendix}%
\begin{center}\Large\bfseries Appendix\end{center}%
\vspace{0.5em}%
}]
\numberwithin{table}{section}
\numberwithin{figure}{section}
\numberwithin{equation}{section}
\section{Particle-Conditioning Robustness}
\label{sec:particle_robustness_appendix}
We vary the particle count on 36 cases with three paired seeds, using the 5,000-step H3 dual-control ablation checkpoint. The initial frame, text, surface depth, camera, and simulated motion are fixed. Table~\ref{tab:particle_conditioning_robustness} reports means over seeds and their variation across counts.
\begin{table}[!htbp]
\centering
\small
\renewcommand{\arraystretch}{1.12}
\setlength{\tabcolsep}{3pt}
\resizebox{\columnwidth}{!}{%
\begin{tabular}{@{}ccccc@{}}
\toprule
Particle count & Native EPE $\downarrow$ & Proxy EPE $\downarrow$
& ROI SSIM $\uparrow$ & Subject $\uparrow$ \\
\midrule
64   & 13.2883 & 6.2604 & 0.8086 & 0.9591 \\
128  & 13.1619 & 6.4210 & 0.8103 & 0.9595 \\
256  & 13.2327 & 6.4848 & 0.8100 & 0.9595 \\
512  & 13.0883 & 6.4407 & 0.8091 & 0.9594 \\
1024 & 13.1542 & 6.5274 & 0.8100 & 0.9588 \\
\midrule
Across-count SD & 0.0771 & 0.1017 & 0.0007 & 0.0003 \\
\bottomrule
\end{tabular}%
}
\caption{Particle-count robustness with fixed depth. The last row is the standard deviation of the five count means.}
\label{tab:particle_conditioning_robustness}
\end{table}

\FloatBarrier
\section{Teacher Evaluation Details}
\label{sec:teacher_metric_details}
\paragraph{Training.}
Each ParForce teacher uses batches of 16 simulated and 16 real depth-only clips and AdamW with weight decay 0.01 (Table~\ref{tab:supp_teacher_train_config}). H3 uses ControlNet with joint text encoding and 90-frame clips at $640\times384$ and 24 fps. The Wan2.1-Fun, Wan2.2-Fun, and CogVideoX-Fun teachers add particle input projections and use 49-frame clips at $384\times256$ and 12 fps. CogVideoX-Fun also learns a first-frame projection.

\begin{table}[H]
\centering\footnotesize
\setlength{\tabcolsep}{3pt}
\begin{tabular}{@{}ccccc@{}}
\toprule
Backbone & Updates & Rank & LoRA LR & Input LR \\
\midrule
H3 33B & 5,000 & 128 & 1 & 1 \\
Wan2.1 1.3B & 4,000 & 512 & 2 & 10 \\
Wan2.2 14B & 4,000 & 128 & 1 & 5 \\
CogVideoX 5B & 4,000 & 128 & 1 & 10 \\
\bottomrule
\end{tabular}
\caption{ParForce teacher training. Learning rates are in units of $10^{-5}$.}
\label{tab:supp_teacher_train_config}
\end{table}

\paragraph{Inference.}
All methods share the first frame, text, and case seed (Table~\ref{tab:supp_teacher_infer_config}). ParForce uses PSS and SDS. Tora and Go-with-the-Track use projected tracks, Go-with-the-Flow uses flow-warped noise, and RealWonder uses simulation RGB and flow-warped noise. Original H3 uses the image and text. Scoring uses the first three seconds of physical motion at $384\times256$.

\begin{table}[H]
\centering\footnotesize
\setlength{\tabcolsep}{2.5pt}
\begin{tabular}{@{}ccccccc@{}}
\toprule
Method & Size & Frames & Hz & Steps & Guide & Shift \\
\midrule
ParForce H3 & $576\times384$ & 90 & 24 & 50 & 1 & 12 \\
Original H3 & $576\times384$ & 90 & 24 & 50 & 1 & 12 \\
ParForce Wan2.1 & $384\times256$ & 49 & 12 & 50 & 5 & 5 \\
ParForce Wan2.2 & $384\times256$ & 49 & 12 & 50 & 5 & 5 \\
ParForce CogVideoX & $384\times256$ & 49 & 12 & 50 & 6 & --- \\
Tora & $720\times480$ & 49 & 12 & 50 & 6 & --- \\
Go-with-the-Flow & $720\times480$ & 49 & 12 & 50 & 6 & --- \\
RealWonder & $832\times480$ & 81 & 20 & 4$^\dagger$ & --- & --- \\
Go-with-the-Track & $832\times480$ & 49 & 12 & 50 & 3 & 5 \\
\bottomrule
\end{tabular}
\caption{Teacher inference settings on the 150 paired cases. Hz is the physical rate. $^\dagger$Steps per autoregressive block.}
\label{tab:supp_teacher_infer_config}
\end{table}

\paragraph{Metrics.}
FID~\cite{heusel2017fid} pools eight Inception frames per clip; FVD~\cite{unterthiner2019fvd} uses 24 post-input frames sampled at 8\,Hz. RAFT EPE~\cite{teed2020raft} compares twelve quarter-second flow pairs under a common mask derived only from ground-truth (GT) motion and forward--backward consistency. Canny F1 uses a $5\times5$ Gaussian filter ($\sigma=1$), thresholds 50/100, and a two-pixel matching tolerance. Relative-depth nRMSE uses Depth Anything V2-Small~\cite{yang2024depthanythingv2}, jointly fits one positive scale and offset across the eight scoring frames, and normalizes by the GT depth's 95th--5th percentile range. The VLM judge receives the first frame, instruction, and sixteen generated frames, without method identity or future RGB. Its visual-integrity and physical-plausibility columns report the fraction scoring at least eight out of ten.

\FloatBarrier
\section{Augmented Physical Condition Constructor}
\label{supp:constructor}

The constructor takes an initial RGB frame and a prompt with an initial-scene description and an action instruction. It produces two synchronized controls for the teacher, a perspective particle state sequence (PSS) and a surface-depth sequence (SDS). Figure~\ref{fig:supp_constructor_pipeline} shows the construction stages.

\begin{figure}[!ht]
\centering
\footnotesize
\begingroup
\setlength{\tabcolsep}{4pt}
\renewcommand{\arraystretch}{1.11}
\begin{tabular}{@{}p{0.28\columnwidth}p{0.67\columnwidth}@{}}
\rowcolor{black!8}
\multicolumn{2}{l}{\textbf{Input} \quad First RGB frame $+$ initial scene and action} \\
\multicolumn{2}{c}{$\downarrow$} \\
\rowcolor{cvprblue!9}
\textbf{1. Ground} & Identify movers, force targets, supports, and contacts. \\
\textbf{2. Fit} & Estimate masks, camera, and geometry. Check reprojection and support. \\
\rowcolor{cvprblue!9}
\textbf{3. Plan} & Map the action to an impulse, force, actuator, or attachment constraint. \\
\textbf{4. Simulate} & Check the initial state and advance the scene with a physics solver. \\
\rowcolor{cvprblue!9}
\textbf{5. Render} & Export PSS and SDS with a common camera and clock. \\
\textbf{6. Validate} & Check the outputs. Repair the earliest failing stage and rerun dependent stages. \\
\multicolumn{2}{c}{$\downarrow$} \\
\rowcolor{black!8}
\multicolumn{2}{l}{\textbf{Output} \quad PSS, SDS, and the validation record} \\
\end{tabular}
\endgroup
\caption{First-frame physical condition construction.}
\label{fig:supp_constructor_pipeline}
\end{figure}

\subsection{Constructor Evaluation}
\label{supp:constructor_eval}
\begin{figure}[H]
    \centering
    \includegraphics[width=\linewidth,height=0.92\textheight,keepaspectratio]{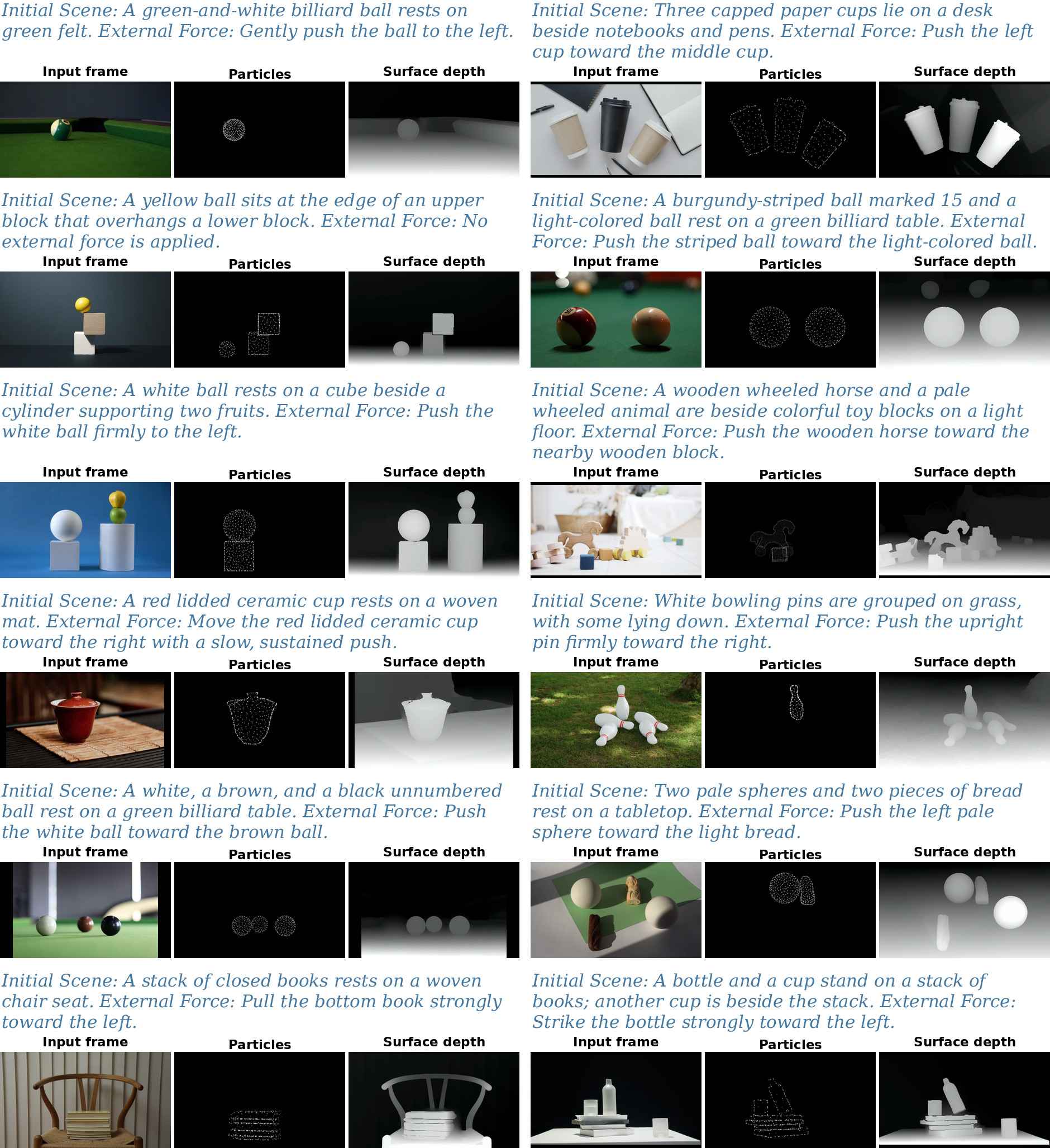}
    \caption{Inputs and reconstructed conditions for 12 cases. Each example shows the prompt, input first frame, and the perspective particles and surface depth at the same simulated time. All 12 cases yielded both conditions. Two failed the automated technical checks.}
    \label{fig:supp_constructor_examples}
\end{figure}

We evaluate 40 real first frames across five scenes. The constructor uses verified SAM~2.1~\cite{ravi2024sam2} and MoGe-1 ViT-L~\cite{wang2025moge} caches, frozen scene roles, Qwen3.5~\cite{qwen35_release} action planning and visual review, and MuJoCo~\cite{todorov2012mujoco} contact dynamics. Figure~\ref{fig:supp_constructor_examples} (Sec.~\ref{supp:constructor_examples}) shows selected particle and depth outputs.

\begin{table}[H]
\centering
\small
\setlength{\tabcolsep}{3.5pt}
\begin{tabular}{lcccc}
\toprule
Scene & $N$ & IoU & Pass & Time (s) \\
\midrule
Stacks          & 7 & 0.7262 & 7/7 & 31.8 \\
Containers      & 9 & 0.8950 & 9/9 & 32.0 \\
Supports        & 7 & 0.9553 & 6/7 & 29.4 \\
Collisions      & 11 & 0.9459 & 8/11 & 30.4 \\
Single object   & 6 & 0.9431 & 6/6 & 27.8 \\
\midrule
All             & 40 & 0.8911 & 36/40 & 30.4 \\
\bottomrule
\end{tabular}
\caption{Constructor results on 40 real first frames.}
\label{tab:supp_constructor}
\end{table}

We compute first-frame IoU for the 26 cases with a valid pairing between projected visible geometry and independently annotated RGB outlines.

\paragraph{Acceptance.}
All 40 cases produced both conditions. A case passes when it satisfies the automated technical checks and all four visual checks. In total, 36 cases passed.

\paragraph{Timing.}
Timing starts with the first-frame segmentation, depth, and verified scene roles ready and ends when the Qwen visual verdict is parsed. For each case, we sum measured durations for Qwen action planning (including response-format retries), geometry fitting, physics simulation, both condition renders and encodes, technical checks, review preview preparation, and Qwen visual review. Table~\ref{tab:supp_constructor} reports the median of these per-case sums, 30.4~s over all 40 cases. First-frame cache generation, manual annotation, model loading, queueing, whole-case reruns, and teacher video inference are outside this window.

\FloatBarrier
\section{Condition-Injection Ablation}
\label{supp:injection}

Our goal is to transfer simulated motion and geometry to video while preserving the appearance of the real input image. We compare three designs for injecting PSS and SDS into H3.

\paragraph{Channel concatenation.}
A learned projection adds concatenated PSS and SDS features to each video token, keeping the sequence length fixed. Attention and feed-forward LoRA in the H3 trunk continue to affect denoising when the condition projection is bypassed.

\paragraph{Token concatenation.}
Each joint PSS/SDS patch becomes an extra token at the corresponding spacetime position. Appending $M$ condition tokens to $N$ existing tokens raises the number of dense attention interactions from $N^2$ to $(N+M)^2$ at every denoising step and adds tokenwise computation throughout the trunk.

\paragraph{ControlNet.}
A separate branch~\cite{h3_fun_union2_release} encodes PSS and SDS and injects residuals at selected H3 layers. Training updates the branch's LoRA, input, and residual projections. The backbone stays frozen and its token sequence length stays fixed.

\subsection{Channel Concatenation vs. ControlNet}

Both variants use the same H3 base, training data, training budget, and inference inputs, including the initial noise. Channel concatenation updates the denoising backbone through LoRA, so bypassing the physical inputs still leaves a 13.21\% deviation from the original H3 prediction. With physical guidance active, it also yields higher dynamic depth error and background LPIPS than ControlNet (Table~\ref{tab:supp_injection}).

\par\smallskip\noindent
\begin{minipage}{\columnwidth}
\captionsetup{type=table}
\centering\footnotesize
\setlength{\tabcolsep}{3.5pt}
\begin{tabular}{@{}cccc@{}}
\toprule
Injection & \shortstack{Base Prediction\\Deviation $\downarrow$\\(Controls Off)} & \shortstack{Dynamic SDS\\nRMSE $\downarrow$} & \shortstack{Background\\LPIPS $\downarrow$} \\
\midrule
Channel concat. & 13.21\% & 0.3320 & 0.2291 \\
Alibaba ControlNet & \textbf{0.00\%} & \textbf{0.2424} & \textbf{0.1859} \\
\bottomrule
\end{tabular}
\caption{Condition-injection comparison. All entries are means.}
\label{tab:supp_injection}
\end{minipage}
\par\smallskip

\emph{Base Prediction Deviation (Controls Off)} measures how closely each trained model matches frozen H3 without physical guidance. With the trained weights loaded, we bypass PSS and SDS and compare the first-step video velocity to that of frozen H3 using the same first frame, text, and initial noise. We divide the $\ell_2$ norm of the difference by the $\ell_2$ norm of the frozen H3 prediction and report the mean percentage over 53 cases.

The other two metrics use guided outputs. Dynamic SDS nRMSE compares the conditioning depth to the estimated output depth~\cite{yang2024depthanythingv2} in the dynamic region after affine alignment, over 53 cases. Background LPIPS~\cite{zhang2018lpips} compares later frames to the input image in a shared static mask, excluding particle coverage and depth changes, over 49 real-image inputs.

\subsection{Cost of Reference-Token Concatenation}

We time H3 Ref2VA~\cite{minimax_h3_release}, which appends a separate token sequence per reference video. The learned variant above packs both modalities into one sequence. We compare one versus two 2-second references with the same first image, text, and noise. Each run generates 124 frames at $640\times384$ in 50 steps, with one case per rank on eight Ascend 910B3 NPUs. We report end-to-end medians over 37 pairs after first-use runs (Table~\ref{tab:supp_token_latency}).

\par\smallskip\noindent
\begin{minipage}{\columnwidth}
\captionsetup{type=table}
\centering\footnotesize
\setlength{\tabcolsep}{5pt}
\begin{tabular}{ccc}
\toprule
Physical references & Latency (s) & Relative latency \\
\midrule
SDS & 1,571.4 & $1.00\times$ \\
PSS $+$ SDS & 2,579.4 & $1.64\times$ \\
\bottomrule
\end{tabular}
\caption{H3 Ref2VA latency from input processing to video export. Relative latency is the median paired ratio.}
\label{tab:supp_token_latency}
\end{minipage}
\par\smallskip

\FloatBarrier
\section{AP-DMD Implementation}
\label{supp:dmd}

\paragraph{Models and conditioning.}
The generator $G_\theta$ starts from H3 Turbo's released four-step LoRA~\cite{lightx2v_h3_turbo2026}. The fake-score model $F_\psi$ starts from base H3 with a fresh LoRA. The teacher $T_\phi$ is the frozen 5,000-step H3 checkpoint.

All three models share joint image--text features for $c=(I_0,s,a)$ and the first-frame prefix. The teacher also receives fixed controls $h=(\mathbf P,\mathbf D)$, with a zero PSS latent for depth-only clips. Losses cover generated tokens.

\paragraph{Four-step rollout.}
Starting from Gaussian noise, $G_\theta$ takes four Euler steps using the shifted grid
\begin{equation}
 u_k=1-k/4,\qquad \sigma_k=\frac{6u_k}{1+5u_k},
 \quad k=0,\ldots,4.
\end{equation}
At each step,
\begin{equation}
 x_{k+1}=x_k+(\sigma_k-\sigma_{k+1})
 v_\theta(x_k,\sigma_k;c).
\end{equation}
H3 takes $1-\sigma_k$ and predicts clean-minus-noise velocity. We apply the endpoint gradient at $y=x_4$ through one uniformly selected denoising step using a straight-through approximation.

\paragraph{Generator update.}
Draw $b\sim U(0.02,0.98)$ and set $\sigma=6b/(1+5b)$. With fresh Gaussian $\epsilon$, score $x=(1-\sigma)\operatorname{sg}(y)+\sigma\epsilon$ at the same $\sigma$ with both networks:
\begin{equation}
 \hat z_T=x+\sigma v_\phi(x,\sigma;c,h),\quad
 \hat z_F=x+\sigma v_\psi(x,\sigma;c).
\end{equation}
Here $\operatorname{sg}$ denotes stop-gradient. The teacher and fake-score predictions are computed without gradients. The generator uses the per-sample direction
\begin{align}
 m&=\max\bigl(\operatorname{mean}|\operatorname{sg}(y)-\hat z_T|,0.01\bigr),\\
 g&=(\hat z_F-\hat z_T)/m,\\
 \mathcal L_G&=\tfrac12\operatorname{mean}
        \bigl(y-\operatorname{sg}(y-g)\bigr)^2 .
\end{align}
The mean is over generated latent dimensions. The detached target applies the normalized score-difference direction directly, omitting the reverse-KL derivation's time multiplier~\cite{yin2024onestepdiffusiondistributionmatching,yin2024improveddistributionmatchingdistillation}.

\begin{table}[H]
\centering\small
\setlength{\tabcolsep}{4pt}
\begin{tabular}{@{}lp{0.43\columnwidth}@{}}
\toprule
Setting & Value \\
\midrule
Generator / fake-score updates & 200 / 1,000 \\
LoRA rank and alpha & 128 for both networks \\
Generator / fake-score LR & $3\times10^{-6}$ / $10^{-5}$ \\
Optimizer & AdamW, $\beta=(0,0.999)$ \\
Weight decay / grad norm cap & 0.01 / 1.0 \\
Global batch / per-device batch & 32 / 1 \\
Training clips & $640\times384$, 90 frames, 24 fps \\
Teacher guidance scale & 1.0 \\
Student export interval & 50 generator updates \\
\bottomrule
\end{tabular}
\caption{Optimization settings for the reported AP-DMD checkpoint.}
\label{tab:supp_dmd_config}
\end{table}

\paragraph{Fake-score updates.}
Each generator update is followed by five fake-score updates, with zero fake-only warmup. Each uses fresh data and noise and a detached rollout $y'$. Draw $\sigma'$ from the same shifted distribution and set $x'=(1-\sigma')y'+\sigma'\epsilon'$ for flow matching~\cite{lipman2023flowmatching}:
\begin{equation}
 \mathcal L_F=\operatorname{mean}
 \bigl(v_\psi(x',\sigma';c)-(y'-\epsilon')\bigr)^2 .
\end{equation}
The generator and fake-score model have separate optimizers; Table~\ref{tab:supp_dmd_config} summarizes the optimization settings.

\FloatBarrier
\section{Additional Results}
\label{supp:student_qualitative}

\paragraph{Physics-IQ Verified Results.}
Models receive the first frame and official prompt~\cite{motamed2026physicsiq,radsch2026physicsiqverified}. Distillation raises the score from 32.64 to 40.00, with gains on 151 of 198 views and 51 of 66 scenarios. The three motion-overlap metrics account for about 60\% of the gain. Spatiotemporal overlap rises from 16.23 to 22.28.

\begin{figure}[H]
    \centering
    \includegraphics[width=\columnwidth]{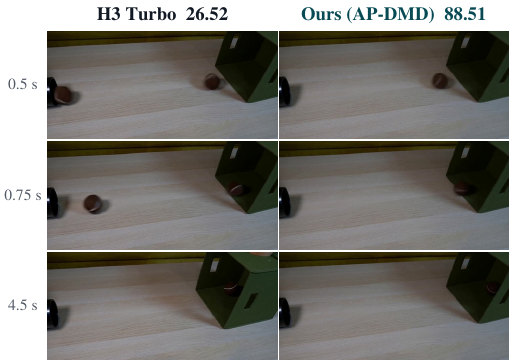}
    \caption{Three matched times for \emph{Roll into Box}. Scores are computed over the full view.}
    \label{fig:supp_roll_into_box}
\end{figure}

\noindent In \emph{Roll into Box} (view 0151), Ours rolls one ball into the box and keeps the box stationary. H3 Turbo generates a second ball near the release tube and later shifts and tilts the box (Fig.~\ref{fig:supp_roll_into_box}). The score rises from 26.52 to 88.51.

\onecolumn
\subsection{Teacher Comparisons}
\begin{figure}[H]
    \centering
    \includegraphics[width=\linewidth,height=0.80\textheight,keepaspectratio]{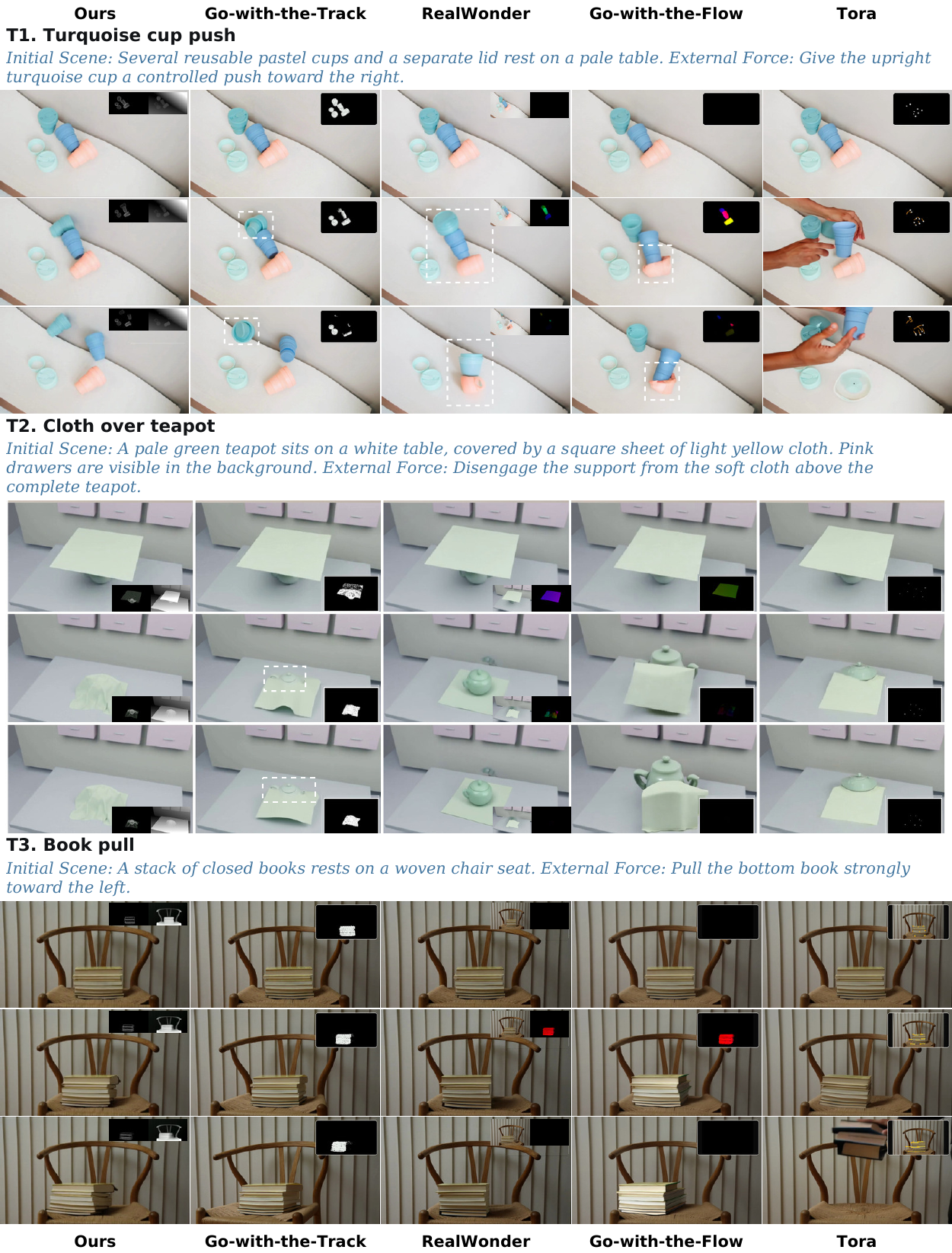}
    \caption{Teacher comparisons for a cup push, cloth released over a teapot, and a book pulled from a stack on a chair. Each block shows three matched times. Paired insets show our particles and surface depth, and RealWonder's simulation render and motion field used for noise warping.}
    \label{fig:supp_teacher_qualitative}
\end{figure}

In \emph{T1 (Turquoise cup push)}, Go-with-the-Track deforms the cup. ParForce preserves its shape and the surrounding layout. In \emph{T2 (Cloth over teapot)}, Go-with-the-Track leaves the top of the teapot exposed. ParForce keeps it covered as the cloth settles. In \emph{T3 (Book pull)}, Go-with-the-Track shifts the chair with the books. ParForce keeps the chair fixed (Fig.~\ref{fig:supp_teacher_qualitative}).

\clearpage
\subsection{Teacher Showcase}
Figure~\ref{fig:supp_teacher_showcase} presents ten additional particle- and depth-guided teacher generation examples across diverse scenes and materials.
\begin{figure}[H]
    \centering
    \includegraphics[page=1,width=\linewidth]{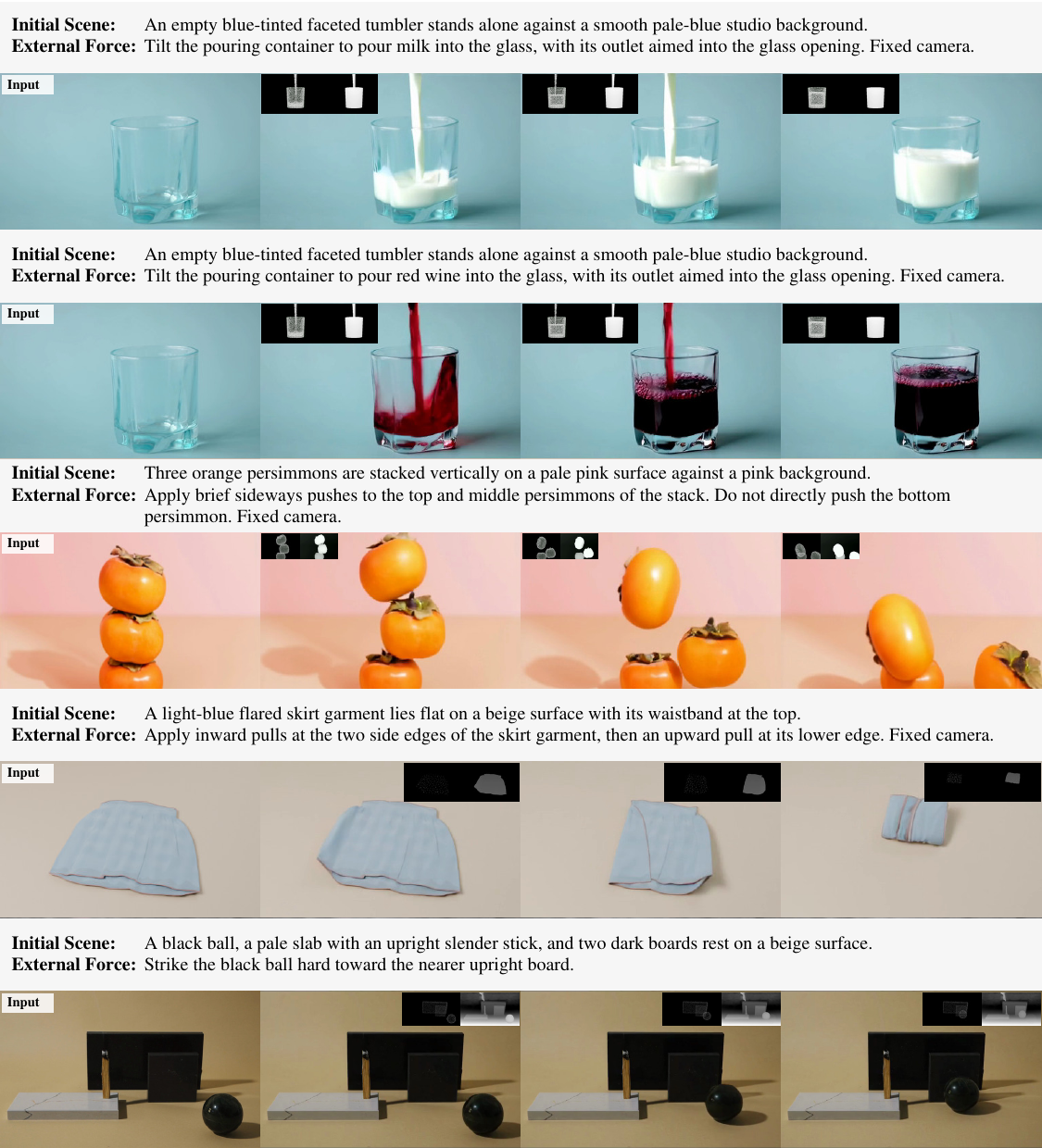}
    \caption{Particle- and depth-guided video generation. Each example shows the input frame and three generated frames, with the particle and surface-depth controls inset.}
    \label{fig:supp_teacher_showcase}
\end{figure}

\clearpage
\subsection*{Teacher Showcase (continued)}
\begin{figure}[H]
    \ContinuedFloat
    \centering
    \includegraphics[page=2,width=\linewidth]{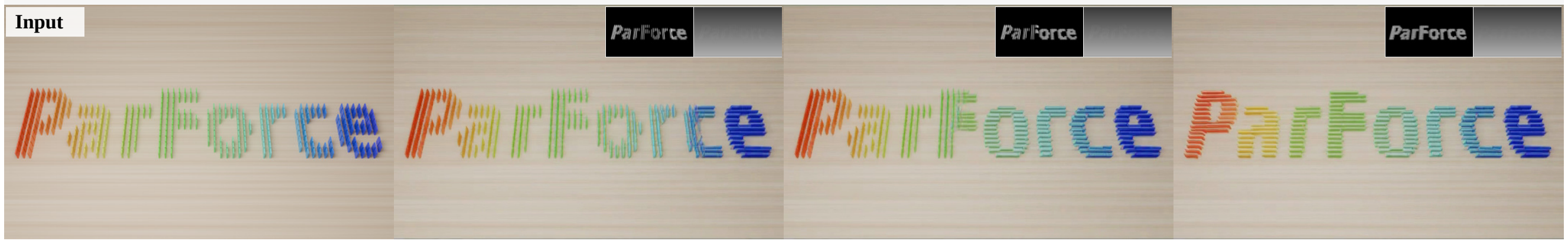}
    \captionsetup{justification=centering}
    \caption{Additional examples shown with the same layout.}
\end{figure}

\clearpage
\subsection{Student Comparisons}
\begin{figure}[H]
    \centering
    \includegraphics[width=\linewidth,height=0.81\textheight,keepaspectratio]{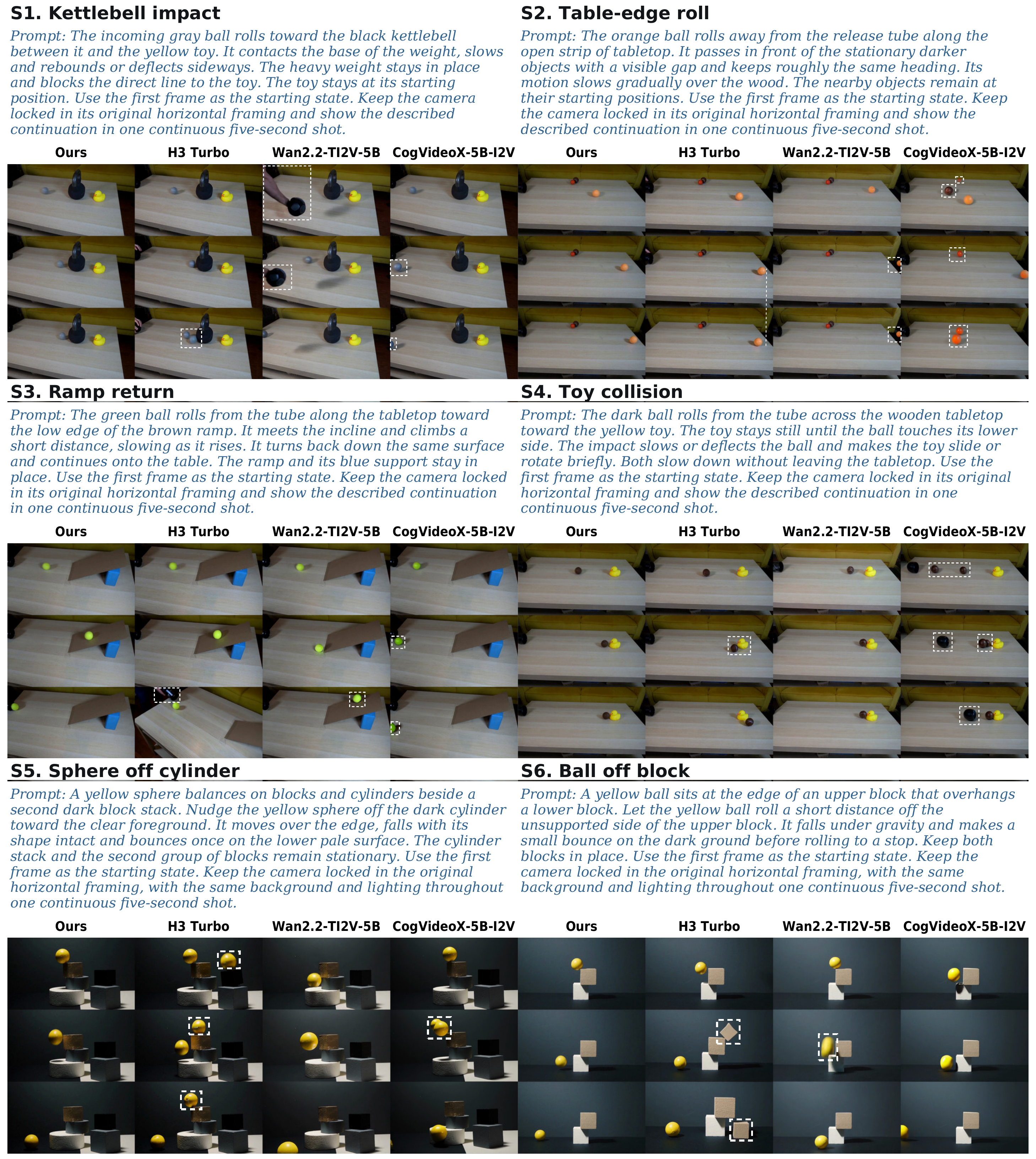}
    \caption{Student comparisons on four Physics-IQ Verified cases and two object-fall scenes. Each example shows three matched playback times. White dashed boxes mark regions of interest.}
    \label{fig:supp_student_physics_iq}
\end{figure}

\paragraph{Qualitative Results with Specified Future Motion.}
A vision-language model (VLM) predicts a continuation from the first frame and initial text. All four generators receive the same frame and continuation prompt (Fig.~\ref{fig:supp_student_physics_iq}). Given this specified future motion, the student renders ball--object interactions more plausibly in S1 and S4 and better preserves object count and support geometry during falls in S5 and S6.

\FloatBarrier
\clearpage
\onecolumn
\section{Constructor Condition Examples}
\label{supp:constructor_examples}

\end{document}